\documentclass[10pt]{article} % For LaTeX2e
\usepackage[accepted]{tmlr}
\usepackage{amsmath,amsfonts,bm}

\def\eqref#1{equation~\ref{#1}}
\def\1{\bm{1}}

\DeclareMathAlphabet{\mathsfit}{\encodingdefault}{\sfdefault}{m}{sl}
\SetMathAlphabet{\mathsfit}{bold}{\encodingdefault}{\sfdefault}{bx}{n}

\DeclareMathOperator*{\argmin}{arg\,min}

\usepackage{hyperref}
\usepackage{url}
\usepackage{wrapfig}
\usepackage{caption}
\usepackage{amsmath}
\usepackage{tabularx}
\usepackage{tikz}
\usepackage{booktabs} % professional-quality tables
\usepackage{amsfonts} % blackboard math symbols
\usepackage{nicefrac} % compact symbols for 1/2, etc.
\usepackage{multirow}
\usepackage{makecell}
\usepackage{graphicx}
\usepackage{caption}
\usepackage{subcaption}
\usepackage{lineno}

\newcommand{\revised}[1]{{#1}} % \textcolor{blue}{#1}}

\title{ProJo4D: Progressive Joint Optimization for Sparse-View Inverse Physics Estimation}

\author{
    \name Daniel Rho \email dnl03c1@cs.unc.edu \\
    \addr University of North Carolina at Chapel Hill
    \AND
    \name Jun Myeong Choi \email chedgekr@cs.unc.edu \\
    \addr University of North Carolina at Chapel Hill
    \AND
    \name Biswadip Dey \email biswa-dey@ieee.org \\
    \addr Meta Reality Labs
    \AND
    \name Roni Sengupta \email ronisen@cs.unc.edu \\
    \addr University of North Carolina at Chapel Hill
}

\def\month{05}  % Insert correct month for camera-ready version
\def\year{2026} % Insert correct year for camera-ready version
\def\openreview{\url{https://openreview.net/forum?id=pqvVrqlXCZ}} % Insert correct link to OpenReview for camera-ready version

\begin{document}

\maketitle

\begin{center}
    \vspace{-15pt}
    \nolinenumbers
    \includegraphics[width=0.85\textwidth]{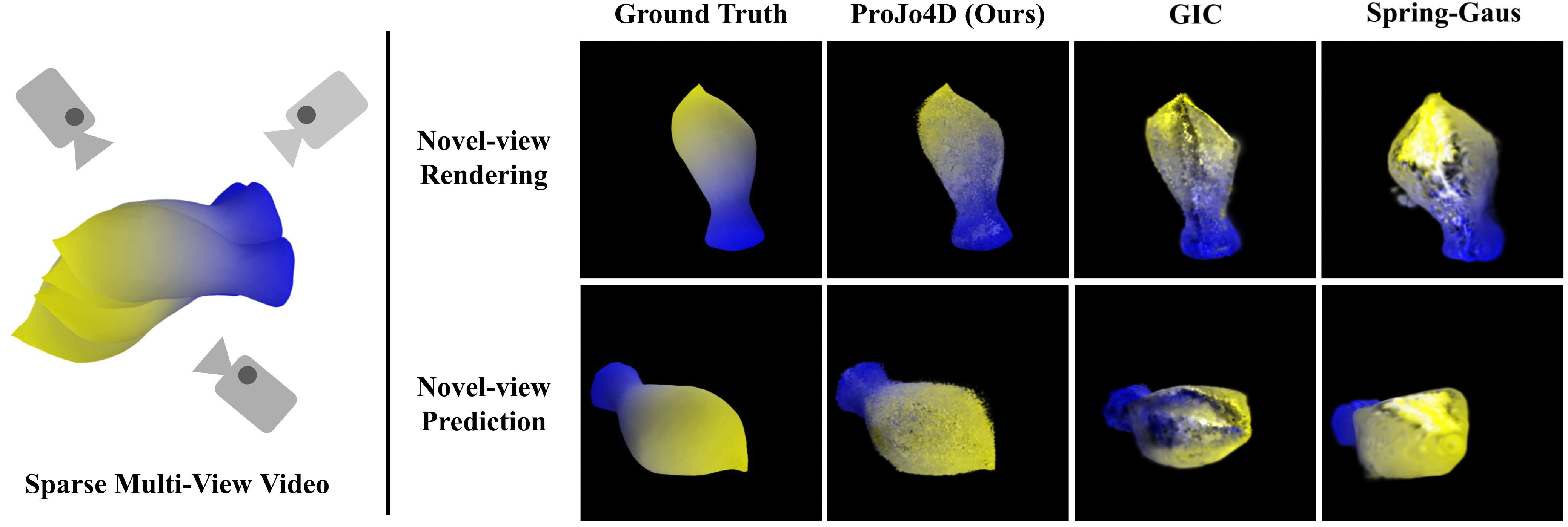}
    \captionof{figure}{
        We present ProJo4D, a progressive joint optimization framework for estimating 4D representation and physical parameters of deformable objects from sparse multi-view video.
        ProJo4D significantly outperforms state-of-the-art inverse physics estimation algorithms, Spring-Gaus \cite{zhong2024springgaus} and GIC \cite{cai2024gic}, which perform sequential optimization of scene geometry and physical parameters.
    }
    \label{fig:teaser}
    \linenumbers
\end{center}

\begin{abstract}
Neural rendering has advanced significantly in 3D reconstruction and novel view synthesis, and integrating physics into these frameworks opens new applications such as physically accurate digital twins for robotics and XR.
However, the inverse problem of estimating physical parameters from visual observations remains challenging.
Existing physics-aware neural rendering methods typically require dense multi-view videos, making them impractical for scalable, real-world deployment.
Under sparse-view settings, the sequential optimization strategies employed by current approaches suffer from severe error accumulation: inaccuracies in initial 3D reconstruction propagate to subsequent stages, degrading physical state and material parameter estimates.
On the other hand, simultaneous optimization of all parameters fails due to the highly non-convex and often non-differentiable nature of the problem.
We propose ProJo4D, a progressive joint optimization framework that gradually expands the set of jointly optimized parameters. This design enables physics-informed gradients to refine geometry while avoiding the instability of direct joint optimization over all parameters.
Evaluations on synthetic and real-world datasets demonstrate that ProJo4D substantially outperforms prior work in 4D future state prediction and physical parameter estimation, achieving up to 10$\times$ improvement in geometric accuracy while maintaining computational efficiency.
\end{abstract}

\section{Introduction}
\label{sec:intro}

Neural rendering techniques have made significant progress in 3D scene reconstruction and novel view synthesis~\cite{nerf,instantngp,3dgs}, but they often lack adherence to the underlying physical laws (e.g., conservation of energy, momentum, or monotonicity constraints).
This gap severely restricts their usage in downstream applications that require not only photorealistic appearance but also physically plausible behavior~\cite{physics_neural_smoke}.
For instance, in vision-based robot learning~\cite{li2024robogsimreal2sim2realroboticgaussian,jiang2025phystwin,abou-chakra2024physically}, agents trained in simulations must seamlessly transfer the learned skills to the real world, which requires accurate physical interactions within the synthetic environment.
Similarly, XR applications in many engineering and industrial settings require rendered objects to respond meaningfully to user interactions (e.g., changes in material properties or object dimensions), environmental constraints, and external forces to maintain immersion, usability, and seamless integration of virtual and physical worlds~\cite{vr-gs,physavatar}.

A recent body of work~\cite{cai2024gic,li2023pacnerf,zhong2024springgaus,Vid2Sim} has attempted to bridge this gap by incorporating physics-based priors into neural rendering pipelines.
However, the state-of-the-art approaches typically rely on dense multi-view setups, often requiring more than ten synchronized cameras with known poses.
Such instrumentation imposes significant practical barriers, particularly in scenarios demanding scalable, flexible, or in-situ data collection.
Whether in robot learning or industrial XR applications, the ability to create physically plausible digital twins from sparse observations is critical for real-world deployment.
Overcoming the dependency on dense multi-view capture is thus crucial to realizing the full potential of neural rendering in physically grounded, deployable systems.

Sparse-view settings pose significant challenges for accurate 3D reconstruction and physical property estimation due to occlusions, shape ambiguities, and limited viewpoints. Existing methods that excel under dense observations degrade markedly when faced with sparse inputs, primarily due to the accumulation of errors in their sequential optimization pipelines~\cite{li2023pacnerf,zhang2024physdreamer,huang2024dreamphysics,zhong2024springgaus,cai2024gic,liu2025physflow}.
These sequential optimization pipelines typically begin by learning an initial 3D or 4D scene representation from sparse images, which is often noisy and ambiguous, particularly in estimating geometry or particle positions.
This flawed representation then serves as the basis for inferring initial physical states (e.g., initial velocities) and subsequently material properties (e.g., stiffness, Poisson's ratio).
As a result, errors introduced early propagate and compound, ultimately degrading both physical state and material parameter estimates.
Although some recent works~\cite{zhong2024springgaus} have explored partial joint optimization of certain parameter subsets, they fall short of addressing the complete inverse physics problem from the outset.
Fully joint optimization of all parameters remains challenging due to the highly non-convex, partly non-differentiable nature of the problem~\cite{zhong2021extending}, often leading to poor local minima, particularly under sparse views.

While prior work has focused on improving scene representations and physical models, the optimization strategy itself remains underexplored despite its critical role in sparse-view settings.
Unlike sequential optimization strategies that optimize parameters one stage at a time, our progressive joint optimization gradually expands the set of jointly optimized parameters.
This maintains coupling between geometry and physical parameters throughout optimization, preventing error accumulation while managing the non-convex optimization landscape.
Progressive optimization succeeds where both alternatives fail. Unlike sequential methods, it allows physics-informed gradients to correct geometry errors throughout training. Unlike full joint optimization, it avoids the instability that arises from optimizing all parameters simultaneously on non-convex landscapes.
As demonstrated in our experiments (Sec.~\ref{sec:exp}), this progressive strategy achieves improved performance compared to both fully sequential and fully joint approaches across diverse materials and sparse-view settings.

To demonstrate the effectiveness of our progressive joint optimization strategy, without any model changes, we utilize GIC's 4D scene representation~\cite{cai2024gic} and physical models, focusing only on changing the optimization strategy. Through extensive evaluations, we show that our progressive joint optimization strategy significantly improves performance for inverse physics estimation on both synthetic and real-world datasets, mitigating the drastic performance drop in sparse-view scenarios.
Our method outperforms the sequential baseline GIC~\cite{cai2024gic}, achieving strong results in 4D future state prediction (Chamfer Distance 16.11 $\rightarrow$ 1.60), future state rendering (PSNR 17.58 $\rightarrow$ 22.30), and physical parameter estimation (Poisson's ratio MAE 0.23 $\rightarrow$ 0.10, Young's modulus MAE 0.18 $\rightarrow$ 0.09), as shown in Table~\ref{tab:springgaus}.
While recent work MASIV~\cite{MASIV} achieves competitive results through neural constitutive models, this approach requires significantly longer optimization time.
In contrast, ProJo4D uses explicit physical parameters and generally outperforms MASIV while maintaining lower computational requirements and physically interpretable parameters crucial for downstream applications.

To summarize, our contributions are as follows.
\begin{itemize}
    \vspace{-8pt}
    \item We identify optimization strategy as a key bottleneck for sparse-view inverse physics estimation, showing that error accumulation in sequential pipelines severely degrades performance.
    \vspace{-8pt}
    \item We propose progressive joint optimization, a simple yet effective strategy that gradually expands the set of jointly optimized parameters, enabling physics-informed gradients to refine geometry while avoiding optimization instability.
    \vspace{-8pt}
    \item We demonstrate that this optimization strategy alone yields up to 10$\times$ improvement in geometric accuracy across multiple datasets, without any architectural modifications.
\end{itemize}

\begin{figure}
    % \vspace{-35pt}
    \centering
    \includegraphics[width=0.9\linewidth]{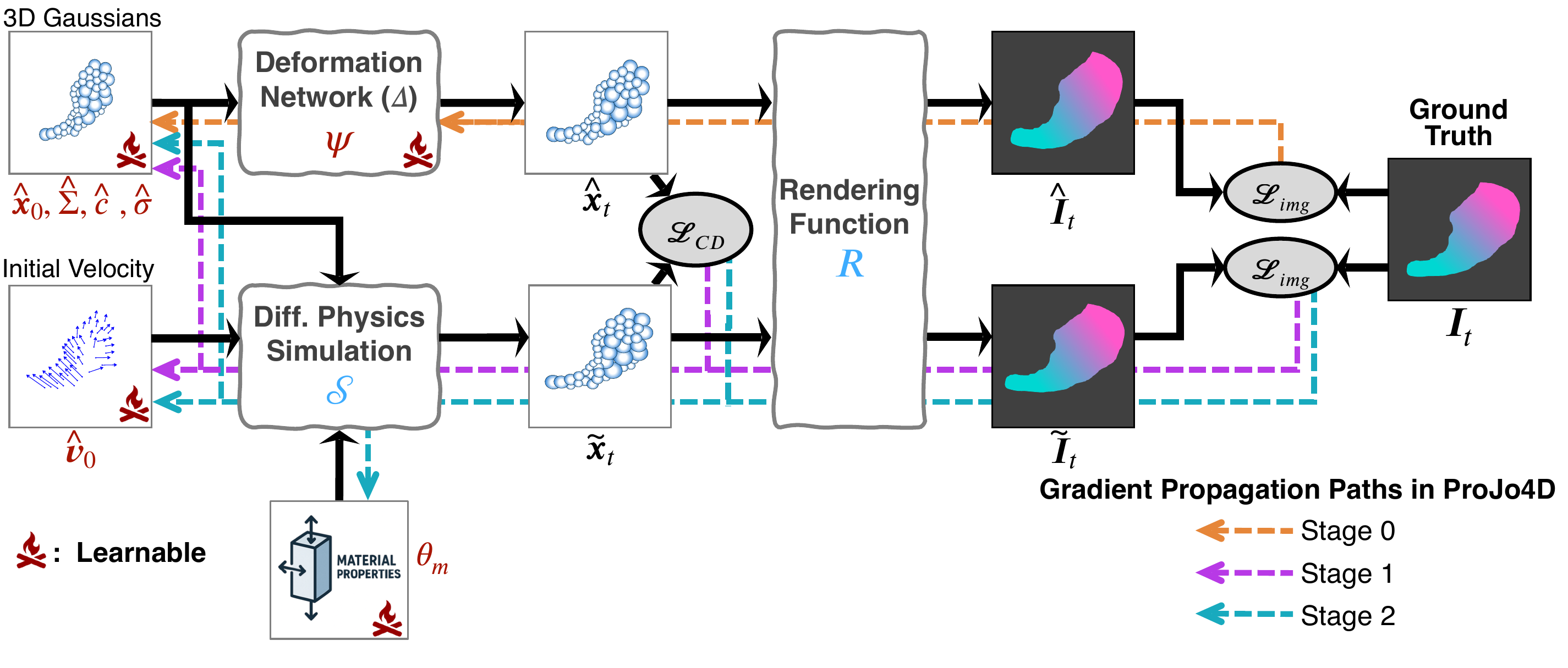}
    \caption{ProJo4D progressively grows the set of optimized variables (3D Gaussian parameters, deformation network, initial velocity, and material properties) across training stages to mitigate error propagation common in sequential frameworks like PAC-NeRF~\cite{li2023pacnerf} and GIC~\citep{cai2024gic}. The diagram illustrates the inter-dependencies among parameters in the inverse physics estimation task, with colored dotted arrows indicating gradient flow during each optimization stage.}
    \label{fig:optimization_overview}
\end{figure}

\section{Background}
\label{sec:background}

\subsection{Notation and Problem Formulation}
\label{ssec:notations}

We introduce the key notations and define the inverse physics estimation problem we aim to solve.

\textbf{Input Data.}
We are given a set of input images $I=\{I_{t,c}\}_{t \in \mathcal{T}, c\in \mathcal{C}}$, where $I_{t,c}$ denotes an image captured at time $t$ from camera $c$.
For each image $I_{t,c}$, the corresponding camera pose $P_c$ from a predefined set of cameras $c \in \mathcal{C}$ and the timestamp $t \in \mathcal{T}$ are assumed to be known. In addition, a transparency alpha map $\alpha_{t, c}$ is often used for initial 3D/4D reconstruction, either rendered or estimated using segmentation or matting.

\textbf{3D / 4D Representation.}
Our scene representation is based on 3D Gaussian Splatting~\citep{3dgs}.
The 3D Gaussians are parameterized by their initial positions $\hat{\mathbf{x}}_0$, covariance matrices $\hat{\Sigma}$, color features $\hat{c}$, and opacity $\hat{\sigma}$.
For representing 4D dynamics, we need additional parameters $\psi$.
The position of a Gaussian at time $t$ is denoted by $\hat{\mathbf{x}}_t$ and is related to its initial position $\hat{\mathbf{x}}_0$ via a displacement function $\Delta(\cdot)$, which models the motion of the Gaussians over time, parameterized by $\psi$ as:
\begin{equation}
    \hat{\mathbf{x}}_t = \hat{\mathbf{x}}_0 + \Delta(\hat{\mathbf{x}}_0, \psi, t).
\end{equation}

We denote the rendering function, a differentiable splatting algorithm~\citep{3dgs}, by $R(\cdot,\cdot,\cdot,\cdot; \cdot)$.
This function takes the Gaussian parameters ($\hat{\mathbf{x}}_t$, $\hat{\Sigma}$, $\hat{c}$, $\hat{\sigma}$) at time $t$ and the camera pose $P_c$ as input, and outputs a rendered image $\hat{I}$:
\begin{equation}
    \hat{I}_{t,c} = R(\hat{\mathbf{x}}_t,\hat{\Sigma},\hat{c},\hat{\sigma}; P_c).
    \label{eq:rendered_image}
\end{equation}
Similarly, $R_{\alpha}(\cdot,\cdot,\cdot; \cdot)$ denotes the alpha map rendering function and $\hat{\alpha}$ denotes the rendered alpha map.
For the detailed rendering process, please refer to 3D Gaussian Splatting~\citep{3dgs}.

\textbf{Physics Parameters.}
Throughout the paper, we refer to both the initial physical state $s$, such as initial velocity $v_0$, and material parameters $\theta_m$ as physical parameters.
Material parameters $\theta_m$ include Young's modulus $E$ and Poisson's ratio $\nu$ for elastic objects.
We assume that the material model, e.g., elastic or plastic, is known a priori, consistent with all other prior works~\citep{li2023pacnerf,cai2024gic}.

A physics simulation model, denoted by $\mathcal{S}(\cdot,\cdot,\cdot,\cdot)$, is used to predict the state of the system over time.
Given the initial positions $\hat{\mathbf{x}}_0$, initial velocity $\hat{\mathbf{v}}_0$, material parameters $\hat{\theta}_m$, the simulation outputs the predicted positions $\tilde{\mathbf{x}}_t$ and the corresponding rendered image $\tilde{I}_{t,c}$ at time $t$ as:
\begin{equation}
    \tilde{\mathbf{x}}_t = \mathcal{S}(\hat{\mathbf{x}}_0, \hat{\mathbf{v}}_0, \hat{\theta}_m, t),
    \quad \tilde{I}_{t,c} = R(\tilde{\mathbf{x}}_t,\hat{\Sigma},\hat{c},\hat{\sigma}; P_c).
\end{equation}

\textbf{Problem Formulation.}
In summary, our task is an inverse estimation problem: given input images $I$, % alpha maps $\alpha$,
their corresponding camera parameters $P$, we aim to estimate the underlying geometry $\textbf{x}_0$, appearance parameters ($\Sigma, c, \sigma$), and physical properties ($v_0$, $\theta_m$) of a deformable object.

\subsection{Related Works}
\label{ssec:physical_param_estimation}
\textbf{Differentiable Physics Simulation.}
Differentiable physics simulation is widely used to optimize and estimate physics-related parameters~\citep{xu2019densephysnet, sanchez2020learning, hudifftaichi, geilinger2020add, zhong2021extending, murthy2020gradsim, NEURIPS2024_510cfd99}.
This forms the foundation of this research area, raising important considerations about which differentiable simulation frameworks to employ and how to design and schedule the optimization process.
Among the commonly used simulation methods are the spring-mass model~\citep{zhong2024springgaus} and the Material Point Method (MPM)~\citep{jiang2016material}.
\revised{MPM is a particle-based method capable of handling diverse deformable materials, such as elastic, plastic, and granular, through different material models.
Differentiable MPM~\citep{hudifftaichi} makes the simulation pipeline differentiable via automatic differentiation, enabling gradient-based optimization of initial positions, velocities, and material parameters.
While MPM itself can simulate multiple materials, existing inverse physics methods~\citep{li2023pacnerf,cai2024gic} typically assume a single material model with global parameters per object.}
Following the same problem setting, we use differentiable MPM as the physics simulator $\mathcal{S}(\cdot,\cdot,\cdot,\cdot)$.
Simplicits~\citep{simplicits} can be used for accelerated inverse physics~\citep{Vid2Sim}, but only supports (hyper)elastic materials with simple gravity-only scenarios.
Our progressive joint optimization shares conceptual similarities with curriculum learning~\citep{bengio2009curriculum} and coarse-to-fine optimization, where easier subproblems are solved before harder ones. However, unlike typical curricula over data or model complexity, we design a curriculum over the parameter space itself.
% Differentiable physics simulation is widely used to optimize and estimate physics-related parameters~\citep{xu2019densephysnet, sanchez2020learning, hudifftaichi, geilinger2020add, zhong2021extending, murthy2020gradsim, NEURIPS2024_510cfd99}.
% This forms the foundation of this research area, raising important considerations about which differentiable simulation frameworks to employ and how to design and schedule the optimization process.
% Among the commonly used simulation methods are the spring-mass model~\citep{zhong2024springgaus} and the Material Point Method (MPM)~\citep{jiang2016material}.
% The gradients from the simulations enable updating physical parameters for system identification.
% Simplicits~\citep{simplicits} can be used for accelerated inverse physics~\citep{Vid2Sim}, but it only supports (hyper)elastic materials with simple gravity-only scenarios.
% We use differentiable MPM as the physics simulator $\mathcal{S}(\cdot,\cdot,\cdot,\cdot)$.
% Our progressive joint optimization shares conceptual similarities with curriculum learning~\citep{bengio2009curriculum} and coarse-to-fine optimization, where easier subproblems are solved before harder ones. However, unlike typical curricula over data or model complexity, we design a curriculum over the parameter space itself.

\textbf{Physics-based Neural Rendering.}
\label{ssec:phys_3d}
Recent neural rendering methods have increasingly incorporated physical priors to enable accurate estimation of scene dynamics and material properties directly from video observations~\citep{li2023nvfi,yu2023hyfluid,Kaneko_2024_CVPR,xue2023dintphys,qiao2022neuphysics,ma2021risp,neurofluid,gao2025seeing}.
These approaches aim to recover physically meaningful parameters, such as material properties, forces, and initial states, by coupling differentiable rendering with physics-based simulation.

Among early attempts, PAC-NeRF~\citep{li2023pacnerf} proposed a general framework that sequentially optimizes geometry and appearance, initial physical states, and material parameters.
Spring-Gaus~\citep{zhong2024springgaus} introduced a spring-mass formulation within the 3D Gaussian Splatting framework~\citep{3dgs} to model dynamic deformations.
Gaussian Informed Continuum (GIC)~\citep{cai2024gic} further improves physical parameter estimation and future-state prediction by leveraging learned 4D representations to guide physically motivated 3D losses.
Vid2Sim~\citep{Vid2Sim} incorporates pretrained models to initialize material parameters and adopts Simplicits~\citep{simplicits} for accelerated per-scene optimization.
MASIV~\citep{MASIV} extends this line by introducing neural material models to replace explicit constitutive laws, achieving material-agnostic representations through learned neural networks~\citep{nclaw}.
While this architectural innovation enables flexibility across material types, it introduces additional model complexity and significantly increases optimization time.
In contrast, our approach maintains the same scene representation and physics framework as prior work~\citep{cai2024gic}, achieving better performance purely through progressive joint optimization without architectural modifications.

Our method focuses on accurate physical parameter estimation from videos under sparse-view settings.
While many existing approaches rely on sequential or stage-wise optimization strategies~\citep{li2023pacnerf,zhong2024springgaus,cai2024gic}, as summarized in Tab.~\ref{tab:sum_opt}, such designs are prone to error accumulation across stages.
In contrast, ProJo4D adopts a progressive joint optimization strategy, which significantly improves robustness and estimation accuracy in sparse-views.

\begin{table}
    % \vspace{-30pt}
    \centering
    \footnotesize
    \setlength{\tabcolsep}{4pt}
    \caption{Optimization strategies of existing methods. X, A, S, and M denote positions, appearances, physical states, and material parameters. 0 denotes initial 3D/4D representation learning before physical parameter estimation. $\triangle$ denotes optional optimization, depending on the scene.}
    \label{tab:sum_opt}
      \begin{tabular}{lcccccc}
        \toprule
        & & & \multicolumn{4}{c}{Stage} \\
        \cmidrule(lr){4-7}
        Method & Param. & 0 & 1 & 2 & 3 & 4 \\
        \midrule
        \multirow{4}*{\begin{tabular}{@{}l@{}}PAC-NeRF~\cite{li2023pacnerf}\\PhysDreamer~\cite{zhang2024physdreamer}\end{tabular}}
         & X & \checkmark & & & & \\
         & A & \checkmark & & & & \\
         & S &  & \checkmark & & & \\
         & M &  & & \checkmark & & \\
        \midrule
        \multirow{4}{*}{Spring-Gaus~\cite{zhong2024springgaus}}
         & X & \checkmark & & & & $\triangle$ \\
         & A & \checkmark & $\triangle$ & & $\triangle$ & $\triangle$ \\
         & S &  & & \checkmark & & $\triangle$ \\
         & M &  & & & & \checkmark \\
        \midrule
        \multirow{4}{*}{\begin{tabular}{@{}l@{}}GIC~\cite{cai2024gic}\\MASIV~\cite{MASIV}\end{tabular}}
         & X & \checkmark & & & & \\
         & A & \checkmark & & & \checkmark & \\
         & S &  & \checkmark & & & \\
         & M &  & & \checkmark & & \\
        \midrule
        \multirow{4}{*}{Vid2Sim~\cite{Vid2Sim}}
         & X & \checkmark & \checkmark & & & \\
         & A & \checkmark & \checkmark & & & \\
         & S &  & & & & \\
         & M & \checkmark & \checkmark & & & \\
        \midrule
        \multirow{4}{*}{Ours}
         & X & \checkmark & \checkmark & \checkmark & &  \\
         & A & \checkmark & \checkmark & \checkmark & &  \\
         & S &  & \checkmark & \checkmark & & \\
         & M &  & & \checkmark & & \\
        \bottomrule
      \end{tabular}
    \vspace{-10pt}
\end{table}

\section{ProJo4D}
\label{sec:method}

Our approach follows a multi-stage pipeline to estimate the constituting parameters: the appearance of a deformable object, initial physical states, and material properties.
As is common in this domain, our pipeline begins with obtaining an initial 3D/4D representation.
Our primary focus lies in the subsequent progressive joint optimization strategy designed to robustly solve inverse physics estimation from limited observations.

\subsection{Motivation}
\label{ssec:motivation}

\textbf{Physical Parameter Estimation.} % Joint Optimization
Estimating the physical state and material parameters of an object from visual observations is a challenging inverse problem.
This difficulty arises primarily because the whole system is non-linear and non-convex.
Moreover, some material models, like non-Newtonian fluids, have non-differentiable but material-parameter-dependent branches.
In addition, some of the physical parameters are strongly coupled, making it difficult to disambiguate their individual contributions from visual cues.
All these difficulties necessitate careful design of optimization strategies to improve the chances of converging to a physically plausible and accurate solution.

\begin{figure}[htp]
    % \vspace{-5pt}
    \centering
    \includegraphics[width=0.5\linewidth]{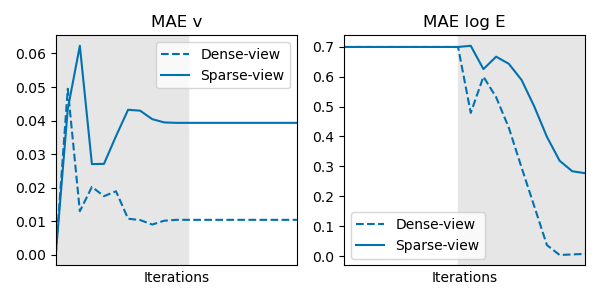}
    \vspace{-10pt}
    \caption{Comparison of error propagation in sequential optimization. % under sparse and dense views.
    The gray region marks the iterations during which the corresponding parameter is optimized: velocity (MAE-$v$, left) and material stiffness (MAE-$\log E$, right).
    Dense views reduce errors faster; sparse views accumulate more errors.
    }
    \label{fig:motivation}
    \vspace{-8pt}
\end{figure}

\textbf{Error accumulation in sparse vs. dense views under sequential optimization.}
Most existing methods rely on sequential optimization~\cite{li2023pacnerf,cai2024gic,zhang2024physdreamer}, where parameters are optimized in stages and estimates from earlier stages are fixed as inputs for later ones. While this strategy can help mitigate some challenges, it introduces a new problem: errors from earlier stages propagate and accumulate, with the effect being substantially worse for sparse-view settings. % compared to dense ones.

For sparse views, the initial geometry estimation is considerably less accurate, and these errors cascade through subsequent optimization stages. This leads to large errors in estimating both initial states and material properties. Figure~\ref{fig:motivation} illustrates this phenomenon by plotting how mean absolute errors (MAE) in velocity ($v$) and material parameters ($\log E$) evolve during the sequential optimization. We exclude the shared initial stage, where the 3D/4D representation is constructed, and focus on the following two stages: velocity optimization (MAE $v$; left) and material parameter optimization (MAE $\log E$; right).
With dense views, errors in the initial 3D/4D representation are smaller, leading to reduced error propagation in subsequent stages, whereas sparse views suffer from greater error accumulation across stages.
For real-world deployments, capturing dense multi-view data with precisely calibrated and synchronized cameras is often impractical. Consequently, mitigating error accumulation and propagation becomes essential to extend the applicability of physics-based 4D reconstruction methods in real-world scenarios.

% \begin{figure}
%   \centering

%   % First subfigure
%   \begin{subfigure}{0.4\linewidth}
%     \centering
%     \includegraphics[width=\linewidth]{figures/fig2b_0.png}
%     \caption{Elastic material model}
%     \label{fig:subfig_a}
%   \end{subfigure}

%   % Second subfigure
%   \begin{subfigure}{0.4\linewidth}
%     \centering
%     \includegraphics[width=\linewidth]{figures/fig2b_1.png}
%     \caption{Sand material model}
%     \label{fig:subfig_b}
%   \end{subfigure}

%   \caption{Material parameter estimation and future prediction performances of different optimization strategies in different material models.}
%   \label{fig:other_optim}
% \end{figure}

\begin{figure}[htbp]
  \centering

  % First subfigure - set to slightly less than 0.5 to allow for spacing
  \begin{subfigure}{0.48\linewidth}
    \centering
    \includegraphics[width=\linewidth]{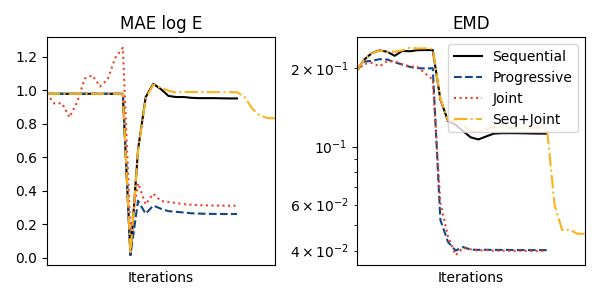}
    \caption{Elastic material model}
    \label{fig:subfig_a}
  \end{subfigure}
  \hfill % This pushes the two images to the far left and right
  % Second subfigure
  \begin{subfigure}{0.48\linewidth}
    \centering
    \includegraphics[width=\linewidth]{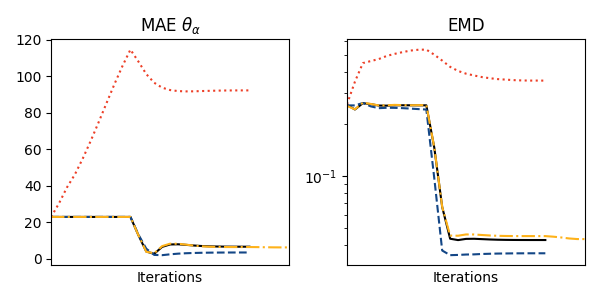}
    \caption{Sand material model}
    \label{fig:subfig_b}
  \end{subfigure}

  \caption{Optimization trajectories for different strategies on elastic (a) and sand (b) materials. Progressive optimization (blue) achieves the best final accuracy while avoiding the instability of joint optimization (red), which diverges for complex materials like sand.}
  \label{fig:other_optim}
\end{figure}

\textbf{Choice of optimization strategies: sequential vs. joint vs. progressive.} 
Figure~\ref{fig:other_optim} illustrates how optimization strategies affect estimation accuracy and robustness across different object shapes and material models. We plot the error in material properties ($\log E$ for elastic (a) and $\theta_{\alpha}$ for sand (b)) and in future 4D state simulation (EMD) over optimization iterations. Sequential optimization, while common, suffers from significant error accumulation as shown in Fig.~\ref{fig:subfig_a}, leading to high estimation errors.

An alternative is joint optimization, adopted by recent approaches such as Vid2Sim~\cite{Vid2Sim}. This strategy can be effective for relatively simple models like elastic objects (Fig.~\ref{fig:subfig_a}), but struggles with more complex systems such as sand (Fig.~\ref{fig:subfig_b}) or non-Newtonian materials, where the optimization landscape is highly non-convex.
In such cases, while initial geometry may already be close to ground truth, the physical parameters are typically far from accurate, causing optimization to stagnate in sub-optimal regions.
% In such cases, initial states may be estimated reasonably well, but material parameters remain inaccurate, causing optimization to stagnate in sub-optimal regions.
A hybrid variant that performs joint optimization after sequential optimization attempts to address this but still inherits the limitations of the initial sequential stage.
% For example, when sequential optimization fails for elastic objects, subsequent joint refinement yields only limited improvements compared to alternatives  (Fig.~\ref{fig:subfig_a}).

These findings highlight the need for more principled optimization strategies that generalize across diverse object shapes and material models.
In the following section, we introduce our progressive joint optimization approach, which mitigates these issues, % and consistently reduces estimation errors,
improving both performance and robustness.

\subsection{Progressive Joint Optimization}

To isolate the effect of our optimization strategy, we adopt the same 4D scene representation (3D Gaussian Splatting with deformation networks) and physics simulation framework (MPM) as GIC~\cite{cai2024gic}.
Unlike recent approaches that introduce new neural architectures for material modeling~\cite{MASIV}, our contribution lies solely in the progressive joint optimization strategy, which gradually expands the set of jointly optimized parameters across three stages (Fig.~\ref{fig:optimization_overview}).

\textbf{Stage 0: Initial 3D / 4D Representation Learning.}
Most existing pipelines, including ours, start with an initial 3D or 4D representation learning stage.
Some methods~\cite{li2023pacnerf,zhong2024springgaus} learn a static 3D scene from the first image of each camera, while others~\cite{cai2024gic} learn a full 4D representation from the multi-view image sequence.
Our method is based on 4D representation learning to leverage 3D guidance during the following stages.
The parameters for 4D representation are optimized by minimizing rendering losses over all frames and cameras:
\begin{equation}
    \mathcal{L}_{img}(\hat{I}, I) = \lambda_{L1}\mathcal{L}_{L1}(\hat{I}, I) + \lambda_{SSIM}\mathcal{L}_{SSIM}(\hat{I}, I),
\end{equation}
\begin{equation}
    \hat{\mathbf{x}}_0^*,\hat{\Sigma}^*,\hat{c}^*,\hat{\sigma}^*,\psi^*
      = \argmin_{\hat{\mathbf{x}}_0, \hat{\Sigma}, \hat{c}, \hat{\sigma}, \psi}
        \sum_{t\in\mathcal{T}}\sum_{c\in\mathcal{C}}
        \lambda_{img}
        \mathcal{L}_{img}\Bigl(
            \hat{I}_{t,c},
            I_{t,c}\Bigr)
      +\lambda_{\alpha}
        % \sum_{t\in\mathcal{T}}\sum_{c\in\mathcal{C}}
        \mathcal{L}_{L1}\Bigl(
            \hat{\alpha}_{t,c},
            \alpha_{t,c}\Bigr),
\end{equation}
where $\hat{I}_{t,c}$ and $\hat{\alpha}_{t,c}$ denote a rendered image and an alpha map, respectively (Sec.~\ref{ssec:notations}).
$\mathcal{L}_{L1}$ and $\mathcal{L}_{SSIM}$ denote L1 loss and Structural similarity index measure (SSIM) loss, and $\lambda_{L1}$, $\lambda_{SSIM}$ are their corresponding loss weights.

\textbf{Stage 1: Initial Physical State Optimization.}
This is the first stage of our progressive optimization strategy, focusing on estimating initial physical states $s$, such as the initial velocity $\hat{v}_0$.
In this stage, we use the first few frames, following prior works~\cite{li2023pacnerf,zhong2024springgaus,cai2024gic}.
This allows the optimization to focus on estimating initial velocity $\hat{v}_0$ before significant deformation or complex interactions take place, separating the influence of initial motion from material response.
At this stage, Gaussian parameters are also optimized.
As the initial velocity $v_0$ is the only physical state parameter in most existing benchmarks, we only optimize $\hat{v}_0$ by minimizing a combined loss over the first few frames $\mathcal{T}_{k}$:
\begin{equation}
\hat{v}_0^*, \hat{\mathbf{x}}_0^*, \hat{\Sigma}^*, \hat{c}^*, \hat{\sigma}^*
 = \argmin_{\hat{v}_0,\hat{\mathbf{x}}_0,\hat{\Sigma},\hat{c},\hat{\sigma}} \;\lambda_{img}\sum_{t\in\mathcal{T}}\sum_{c\in\mathcal{C}}
    \mathcal{L}_{img}\!\Bigl(
        \tilde{I}_{t,c},\,
        I_{t,c}
    \Bigr) +\,\lambda_{geo}
    \sum_{t\in\mathcal{T}}
    \mathcal{L}_{geo}\!\Bigl(
        \tilde{\mathbf{x}}_t,\hat{\mathbf{x}}_t
    \Bigr),
\label{eq:stage1}
\end{equation}
where $\mathcal{L}_{geo}$ is the bidirectional chamfer distance, which measures the distance to the closest point from both estimated and ground truth points.
$\tilde{\mathbf{x}}_t$ and $\tilde{I}_{t,c}$ are positions from physics simulation $\mathcal{S}(\cdot)$ and their corresponding rendered image (Sec.~\ref{ssec:notations}).
$\hat{\mathbf{x}}_t$ denotes the extracted positions from the learned 4D representation using deformation network $\Delta(\cdot)$, as proposed in GIC~\cite{cai2024gic}.

\textbf{Stage 2: Full Joint Optimization.}
After obtaining an improved estimate for physical states $s$, more specifically $v_0$, this stage progresses to include material parameters $\hat{\theta}_m$ during optimization.
For this stage, we utilize data from all frames.
We use the same optimization objective as the previous stage:
% \vspace{-0.3em}
\begin{equation}
\hat{v}_0^*, \hat{\theta}_m^*, \hat{\mathbf{x}}_0^*, \hat{\Sigma}^*, \hat{c}^*, \hat{\sigma}^*
 = \argmin_{\hat{v}_0,\hat{\theta}_m,\hat{\mathbf{x}}_0,\hat{\Sigma},\hat{c},\hat{\sigma}} \;\lambda_{img}\sum_{t\in\mathcal{T}}\sum_{c\in\mathcal{C}}
    \mathcal{L}_{img}\!\Bigl(
        \tilde{I}_{t,c},\,
        I_{t,c}
    \Bigr) +\,\lambda_{geo}
    \sum_{t\in\mathcal{T}}
    \mathcal{L}_{geo}\!\Bigl(
        \tilde{\mathbf{x}}_t,\hat{\mathbf{x}}_t
    \Bigr).
\label{eq:stage2}
\end{equation}

\begin{table*}[ht!]
    \caption{Evaluation on Spring-Gaus~\cite{zhong2024springgaus} dataset with sparse views (3 views). We measure 3D prediction accuracy of future states using Chamfer Distance (CD) and Earth Mover's Distance (EMD), image rendering quality of future states using PSNR and SSIM, and MAE of Young's modulus $E$ and Poisson's ratio $\nu$. For rendering quality evaluation, we used future images from all cameras.
    }
    \label{tab:springgaus}
    \centering
    \resizebox{1\linewidth}{!}{
        \begin{tabular}{rlrrrrrrrr}
            \toprule
             & \textbf{method} & \textbf{apple} & \textbf{banana} & \textbf{chess} & \textbf{cream} & \textbf{cross} & \textbf{paste} & \textbf{torus} & \textbf{mean} \\
            \midrule
            % ------------------------- CD -------------------------
            \multirow{4}{*}{CD \(\downarrow\)}
            & Spring-Gaus & 12.12 & 51.35 & 3.68 & 2.97 & 40.30 & 73.08 & 15.00 & 26.93 \\
            & GIC & 2.13 & 8.37 & 7.51 & 8.16 & 2.51 & 81.24 & 2.81 & 16.11 \\
            & MASIV & 1.02 & 2.23 & 5.15 & 2.42 & 4.75 & 2.33 & 1.51 & 2.77 \\
            & GIC + ProJo4D & \textbf{0.19} & \textbf{0.12} & \textbf{1.37} & \textbf{1.54} & \textbf{0.38} & \textbf{6.93} & \textbf{0.65} & \textbf{1.60} \\
            \midrule
            % ------------------------- EMD -------------------------
            \multirow{4}{*}{EMD \(\downarrow\)}
            & Spring-Gaus & 0.170 & 0.223 & 0.097 & 0.101 & 0.232 & 0.248 & 0.177 & 0.178 \\
            & GIC & 0.090 & 0.106 & 0.139 & 0.135 & 0.084 & 0.263 & 0.081 & 0.128 \\
            & MASIV & 0.053 & 0.051 & 0.124 & 0.088 & 0.102 & 0.110 & 0.046 & 0.082 \\
            & GIC + ProJo4D & \textbf{0.054} & \textbf{0.024} & \textbf{0.066} & \textbf{0.052} & \textbf{0.031} & \textbf{0.142} & \textbf{0.031} & \textbf{0.057} \\
            \midrule
            % ------------------------- PSNR -------------------------
            \multirow{4}{*}{PSNR \(\uparrow\)}
            & Spring-Gaus & 17.03 & 15.79 & 13.85 & 14.62 & 11.24 & 10.94 & 13.01 & 13.78 \\
            & GIC & 20.52 & 21.84 & 14.87 & 13.93 & 22.51 & 12.41 & 17.00 & 17.58 \\
            & MASIV & 21.77 & 24.07 & 15.05 & 16.74 & 21.67 & 15.94 & 18.64 & 19.13 \\
            & GIC + ProJo4D & \textbf{27.10} & \textbf{28.65} & \textbf{17.96} & \textbf{18.76} & \textbf{28.09} & \textbf{15.20} & \textbf{20.354} & \textbf{22.30} \\
            \midrule
            % ------------------------- SSIM -------------------------
            \multirow{4}{*}{SSIM \(\uparrow\)}
            & Spring-Gaus & 0.790 & 0.825 & 0.792 & 0.796 & 0.819 & 0.737 & 0.831 & 0.799 \\
            & GIC & 0.868 & 0.910 & 0.826 & 0.795 & 0.889 & 0.772 & 0.892 & 0.850 \\
            & MASIV & 0.884 & 0.930 & 0.826 & 0.849 & 0.890 & 0.852 & 0.922 & 0.879 \\
            & GIC + ProJo4D & \textbf{0.930} & \textbf{0.959} & \textbf{0.886} & \textbf{0.885} & \textbf{0.943} & \textbf{0.852} & \textbf{0.933} & \textbf{0.913} \\
            \midrule
            % ------------------------- MAE of E -------------------------
            \multirow{2}{*}{MAE $\log E$ \(\downarrow\)}
            & GIC & 0.1840 & 0.4639 & 0.1807 & 0.0838 & 0.3239 & \textbf{0.1380} & 0.2436 & 0.2311 \\
            & GIC + ProJo4D & \textbf{0.0633} & \textbf{0.1519} & \textbf{0.0326} & \textbf{0.0336} & \textbf{0.0469} & 0.1705 & \textbf{0.2315} & \textbf{0.1043} \\
            \midrule
            % ------------------------- MAE of $\nu$ -------------------------
            \multirow{2}{*}{MAE $\nu$ \(\downarrow\)}
            & GIC & 0.1439 & \textbf{0.1049} & 0.0622 & 0.1407 & 0.0955 & 0.2209 & 0.4851 & 0.1790 \\
            & GIC + ProJo4D & \textbf{0.0817} & 0.2237 & \textbf{0.0222} & \textbf{0.0295} & \textbf{0.0307} & \textbf{0.0928} & \textbf{0.1569} & \textbf{0.0911} \\
            \bottomrule
        \end{tabular}
    }
\end{table*}

\begin{figure}[ht]
\centering
\scriptsize

% Time arrows
\begin{minipage}[c]{0.035\textwidth}
  \hspace{0pt}
\end{minipage}%
\begin{minipage}[c]{0.47\textwidth}
  \centering
  \begin{tikzpicture}
\hspace{-1.2em}%  
    \draw[->, thick] (0,0) -- (6,0) node[anchor=west]{\textbf{Time}};
  \end{tikzpicture}
\end{minipage}%
\hspace{2mm}%
\begin{minipage}[c]{0.47\textwidth}
  \centering
  \begin{tikzpicture}
\hspace{-1em}%  
    \draw[->, thick] (0,0) -- (6,0) node[anchor=west]{\textbf{Time}};
  \end{tikzpicture}
\end{minipage}

% Ground Truth
\begin{minipage}[c]{0.035\textwidth}
  \rotatebox{90}{\textbf{G.T}}
\end{minipage}%
\hspace{-1em}%  
\begin{minipage}[c]{0.49\textwidth}
  \includegraphics[width=0.24\textwidth]{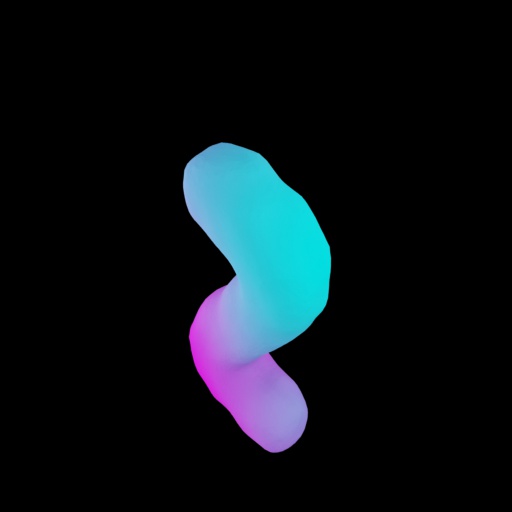}%
\hspace{1mm}%
  \includegraphics[width=0.24\textwidth]{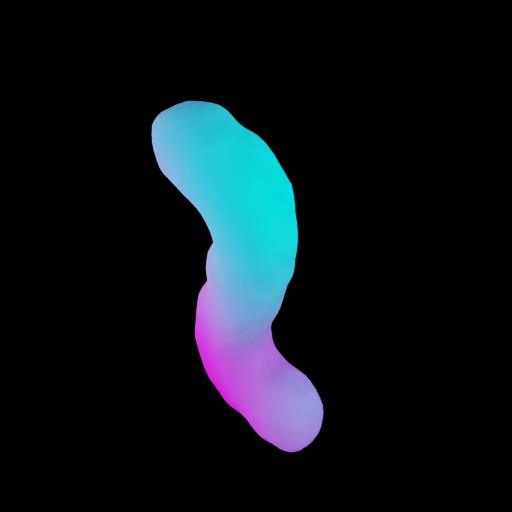}%
  \includegraphics[width=0.24\textwidth]{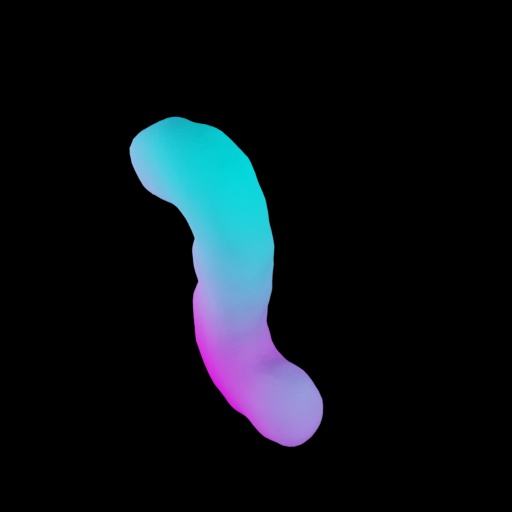}%
  \includegraphics[width=0.24\textwidth]{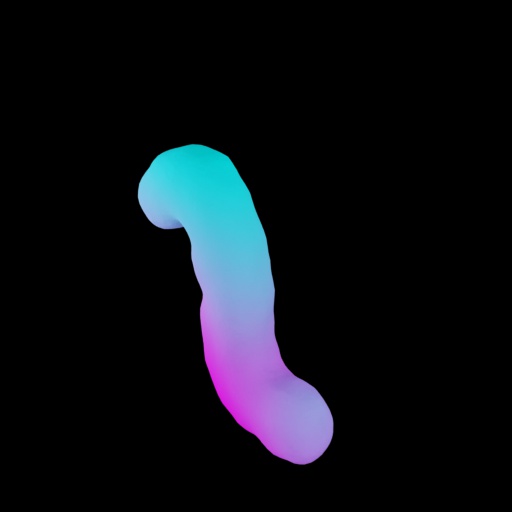}%
\end{minipage}%
\begin{minipage}[c]{0.49\textwidth}
  \includegraphics[width=0.24\textwidth]{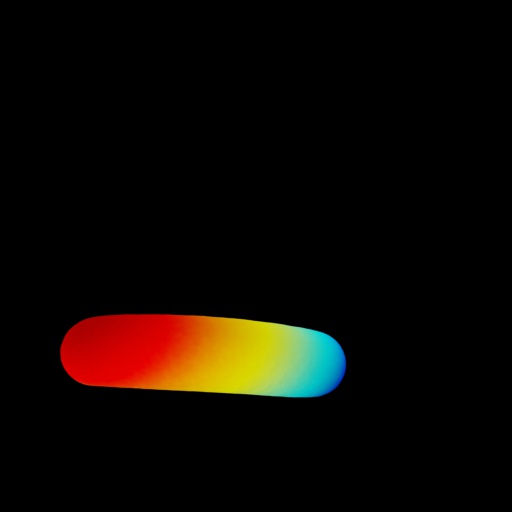}%
\hspace{1mm}%
  \includegraphics[width=0.24\textwidth]{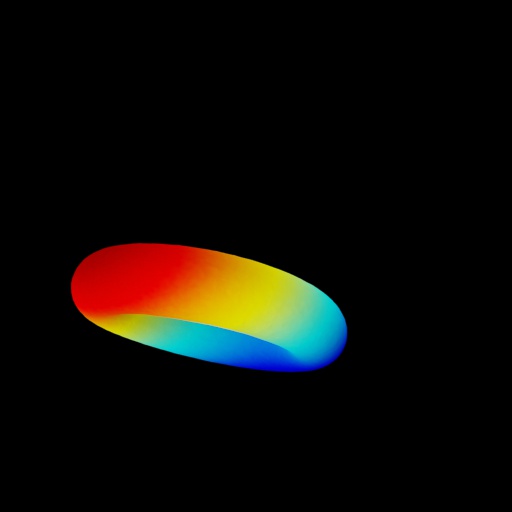}%
  \includegraphics[width=0.24\textwidth]{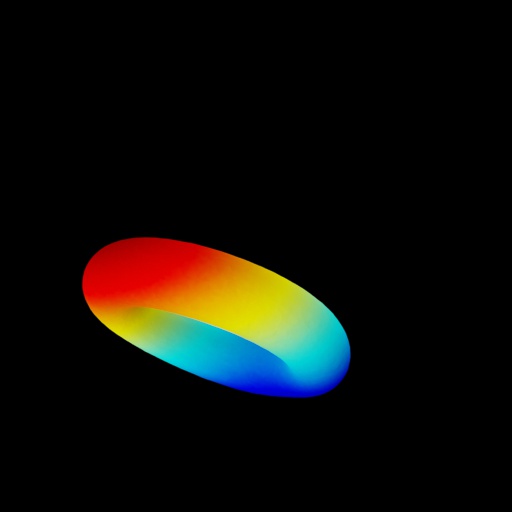}%
  \includegraphics[width=0.24\textwidth]{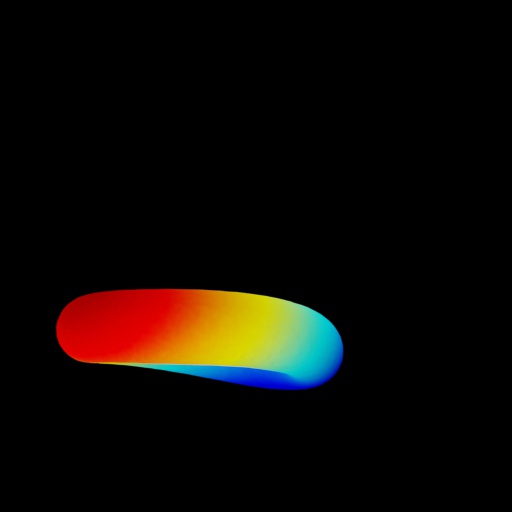}%
\end{minipage}

% GIC
\begin{minipage}[c]{0.035\textwidth}
  \rotatebox{90}{\textbf{GIC}~\cite{cai2024gic}}
\end{minipage}%
\hspace{-1em}%  
\begin{minipage}[c]{0.49\textwidth}
  \includegraphics[width=0.24\textwidth]{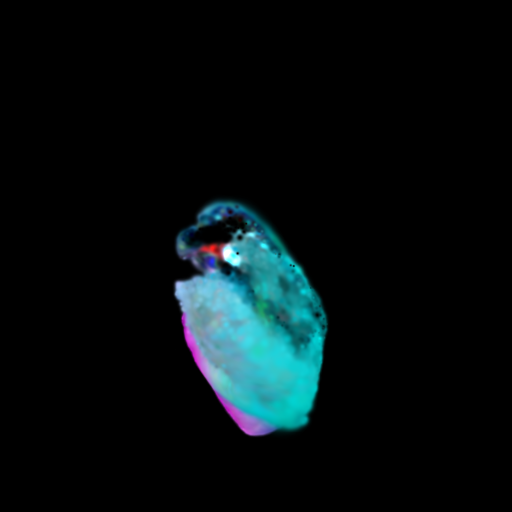}%
\hspace{1mm}%
  \includegraphics[width=0.24\textwidth]{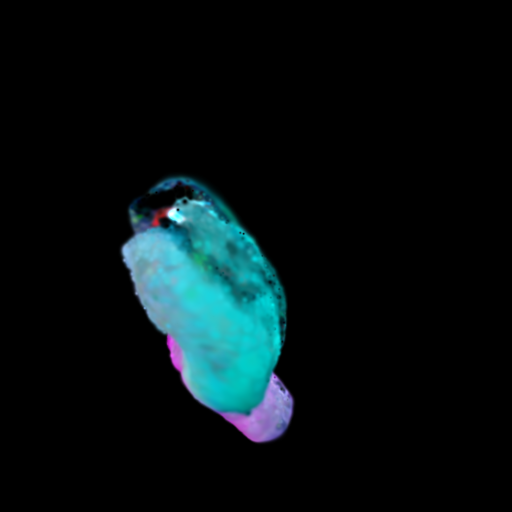}%
  \includegraphics[width=0.24\textwidth]{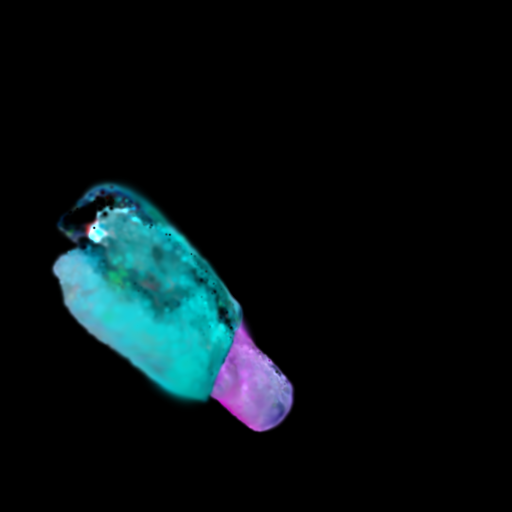}%
  \includegraphics[width=0.24\textwidth]{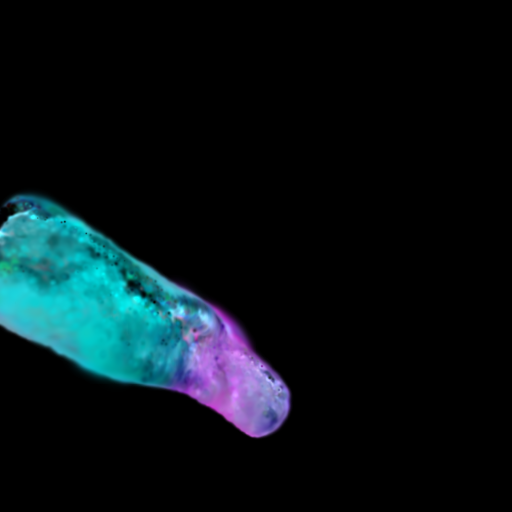}%
\end{minipage}%
% \hspace{2mm}%
\begin{minipage}[c]{0.49\textwidth}
  \includegraphics[width=0.24\textwidth]{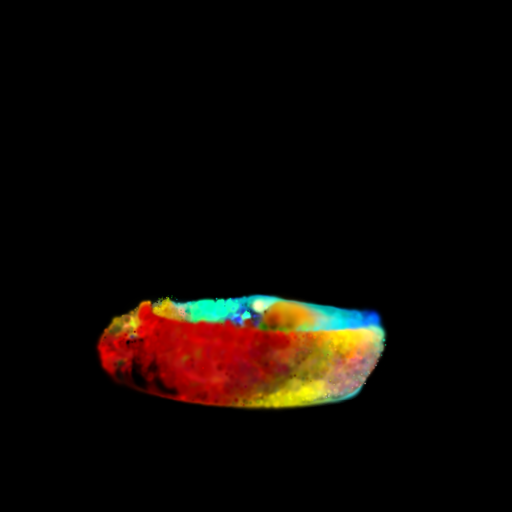}%
\hspace{1mm}%
  \includegraphics[width=0.24\textwidth]{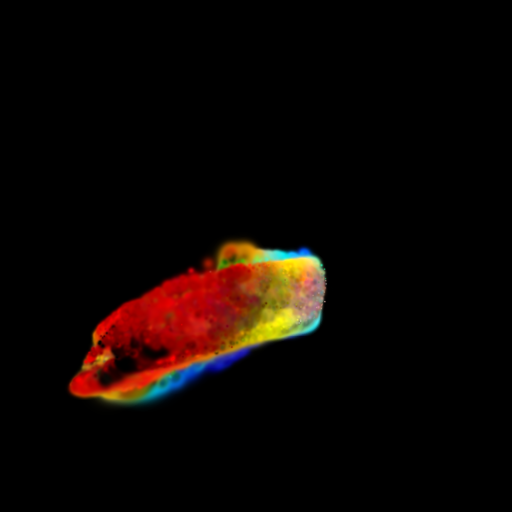}%
  \includegraphics[width=0.24\textwidth]{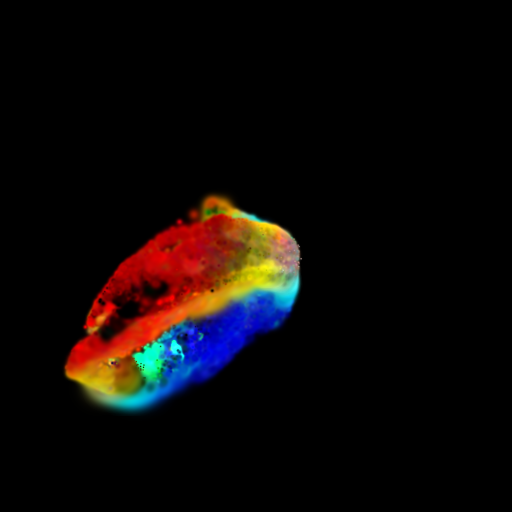}%
  \includegraphics[width=0.24\textwidth]{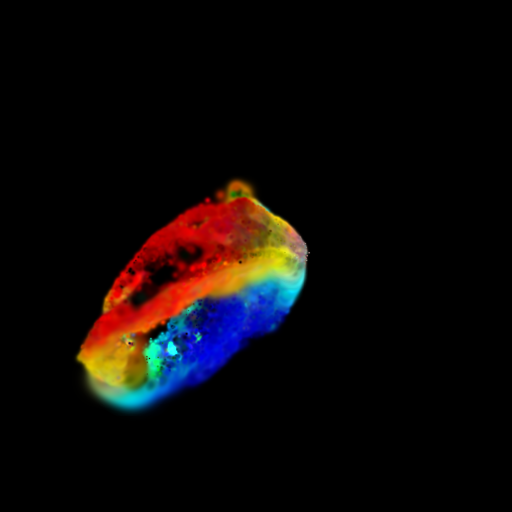}%
\end{minipage}

% Ours
\begin{minipage}[c]{0.035\textwidth}
  \rotatebox{90}{\textbf{Ours}}
\end{minipage}%
\hspace{-1em}%  
\begin{minipage}[c]{0.49\textwidth}
  \includegraphics[width=0.24\textwidth]{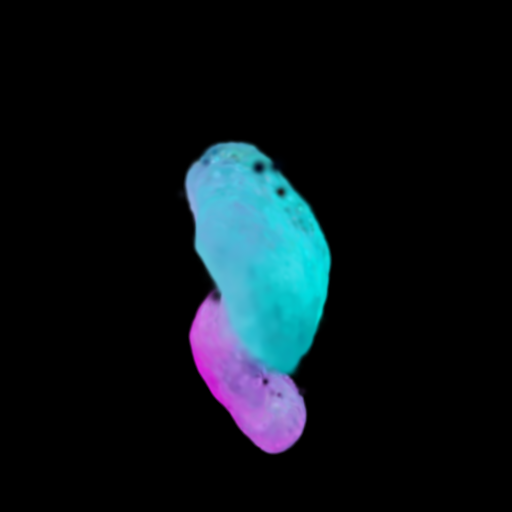}%
\hspace{1mm}%
  \includegraphics[width=0.24\textwidth]{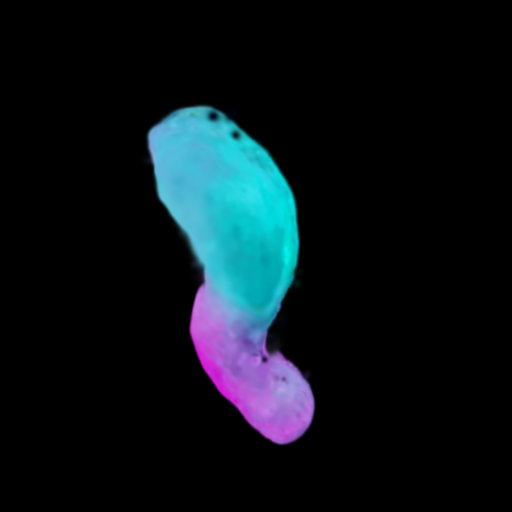}%
  \includegraphics[width=0.24\textwidth]{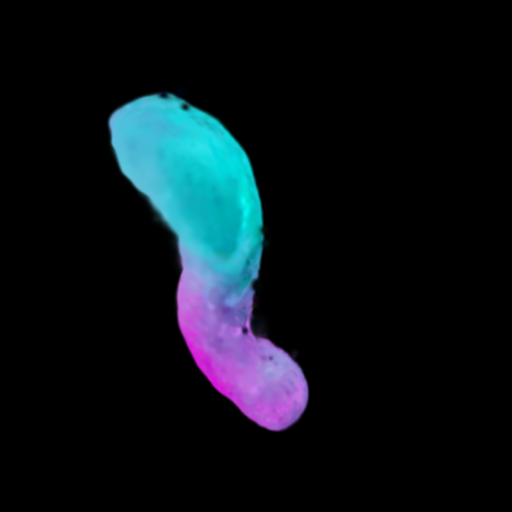}%
  \includegraphics[width=0.24\textwidth]{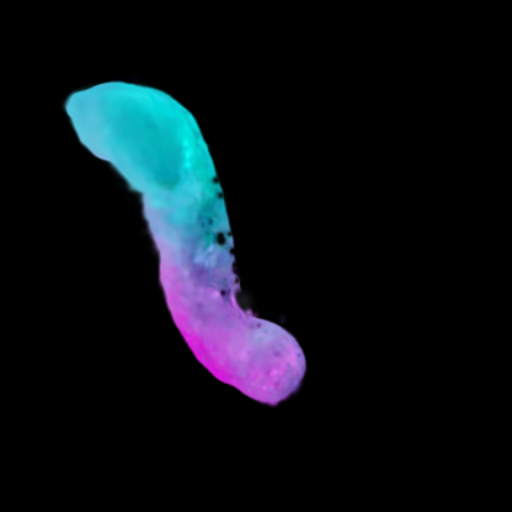}%
\end{minipage}%
% \hspace{2mm}%
\begin{minipage}[c]{0.49\textwidth}
  \includegraphics[width=0.24\textwidth]{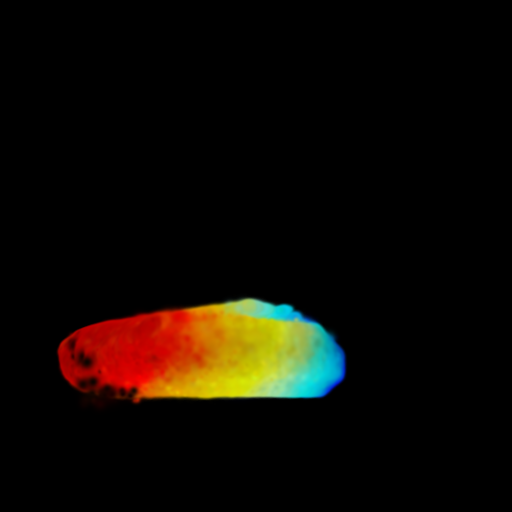}%
\hspace{1mm}%
  \includegraphics[width=0.24\textwidth]{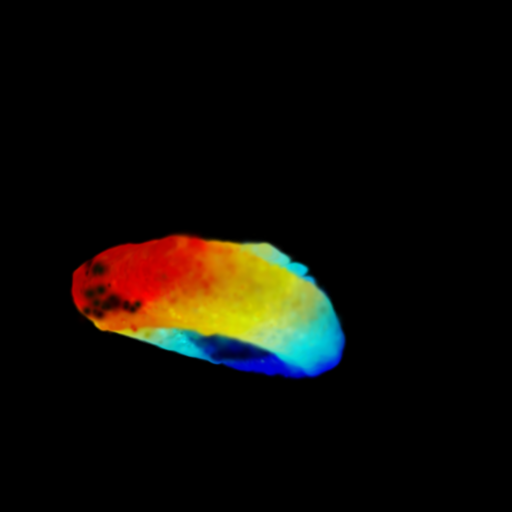}%
  \includegraphics[width=0.24\textwidth]{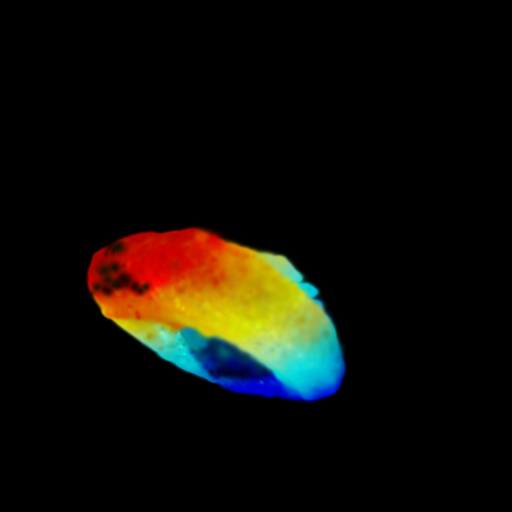}%
  \includegraphics[width=0.24\textwidth]{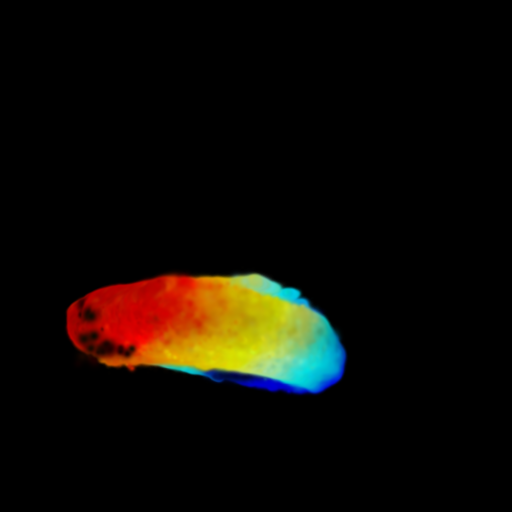}%
\end{minipage}

\hspace{-3mm}%
\begin{minipage}[c]{0.49\textwidth}
\hspace{1em}
  \begin{minipage}[c]{0.23\textwidth}
    \centering
    \textbf{Novel-view\\Rendering}
  \end{minipage}%
  \begin{minipage}[c]{0.75\textwidth}
    \centering
    \textbf{Novel-view Future Prediction}
  \end{minipage}
\end{minipage}
\begin{minipage}[c]{0.49\textwidth}
\hspace{1em}
  \begin{minipage}[c]{0.23\textwidth}
    \centering
    \textbf{Novel-view\\Rendering}
  \end{minipage}%
  \begin{minipage}[c]{0.75\textwidth}
    \centering
    \textbf{Novel-view Future Prediction}
  \end{minipage}
\end{minipage}

\caption{Visual comparison of ProJo4D (Ours) with GIC~\cite{cai2024gic} for novel-view rendering and prediction in future timestep on the Spring-Gaus dataset, using sparse-view inputs. ProJo4D produces more consistent and physically plausible results across both current and future views.}

\label{fig:springgaus}
\end{figure}

\begin{figure}[ht]
\centering
\scriptsize

% Time arrows
% sand: 2
% non-Newtonian: 
\begin{minipage}[c]{0.035\textwidth}
  \hspace{0pt}
\end{minipage}%
\begin{minipage}[c]{0.47\textwidth}
  \centering
  \begin{tikzpicture}
\hspace{-1.2em}%  
    \draw[->, thick] (0,0) -- (5.9,0) node[anchor=west]{\textbf{Time}};
  \end{tikzpicture}
\end{minipage}%
\hspace{2mm}%
\begin{minipage}[c]{0.47\textwidth}
  \centering
  \begin{tikzpicture}
\hspace{-1em}%  
    \draw[->, thick] (0,0) -- (5.9,0) node[anchor=west]{\textbf{Time}};
  \end{tikzpicture}
\end{minipage}

% Ground Truth
\begin{minipage}[c]{0.035\textwidth}
  \rotatebox{90}{\textbf{G.T}}
\end{minipage}%
\hspace{-1em}%  
\begin{minipage}[c]{0.49\textwidth}
  \includegraphics[width=0.24\textwidth]{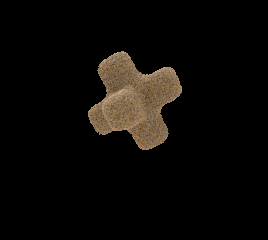}%
  \includegraphics[width=0.24\textwidth]{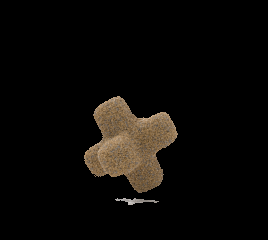}%
  \includegraphics[width=0.24\textwidth]{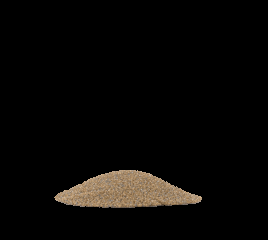}%
  \includegraphics[width=0.24\textwidth]{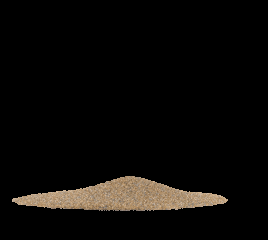}%
\end{minipage}%
\begin{minipage}[c]{0.49\textwidth}
  \includegraphics[trim=130 300 130 0, clip, width=0.24\textwidth]{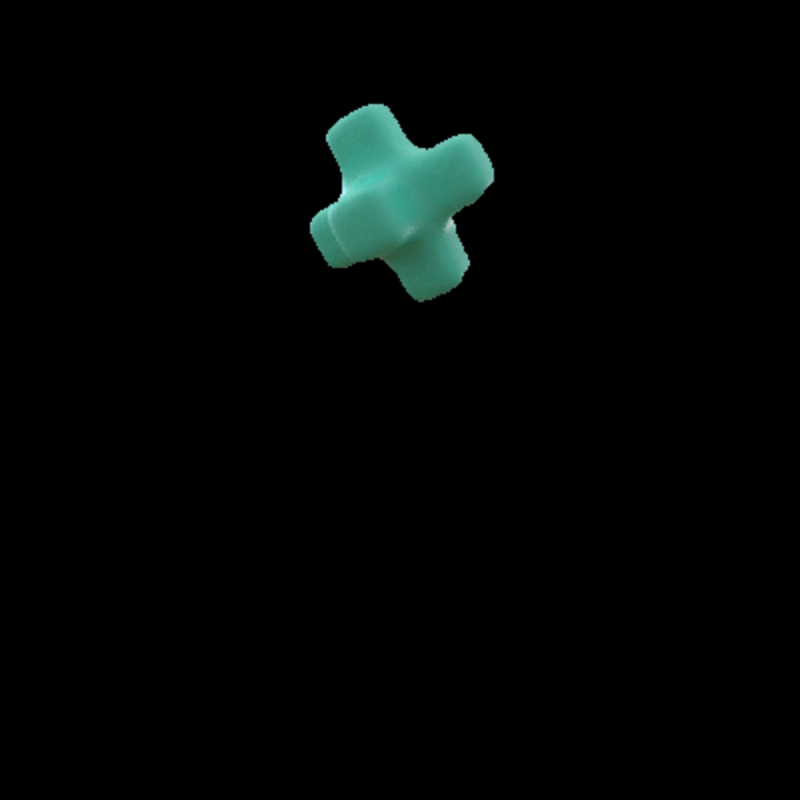}\hspace{-0.275em}
  \includegraphics[trim=130 300 130 0, clip, width=0.24\textwidth]{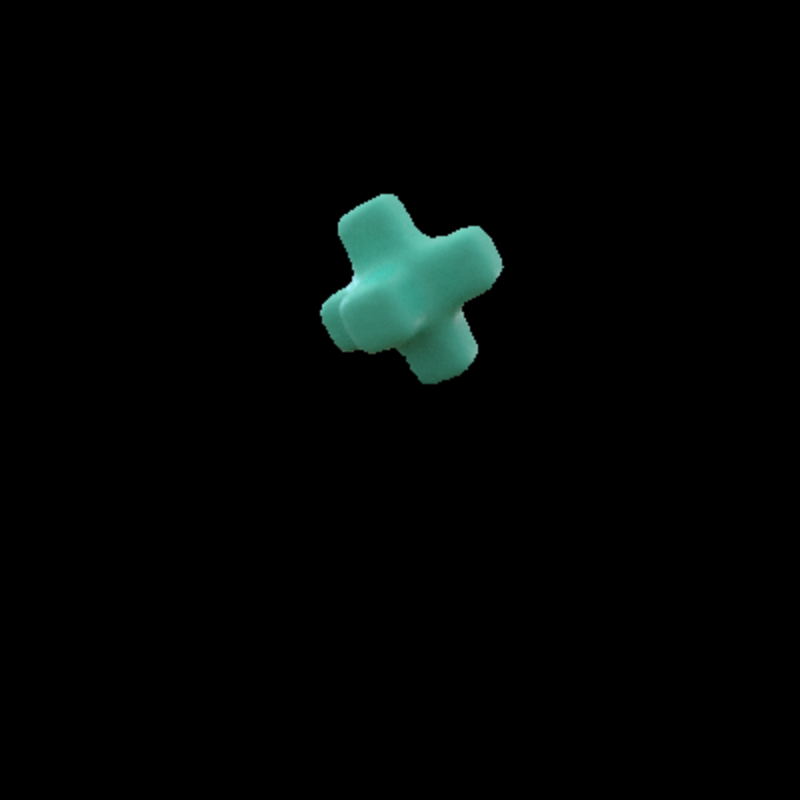}\hspace{-0.275em}
  \includegraphics[trim=130 300 130 0, clip, width=0.24\textwidth]{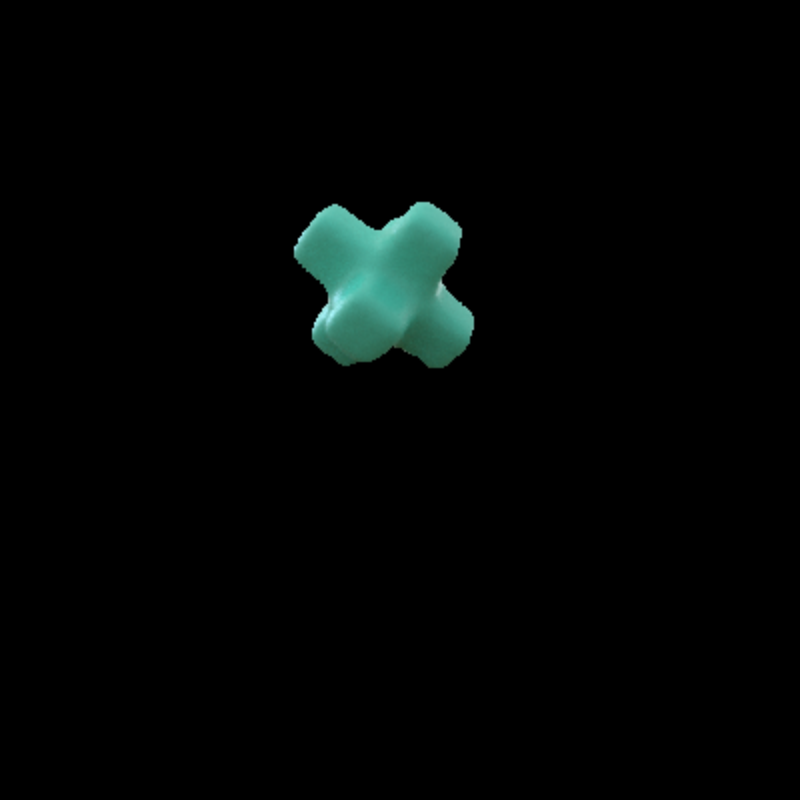}\hspace{-0.275em}
  \includegraphics[trim=130 300 130 0, clip, width=0.24\textwidth]{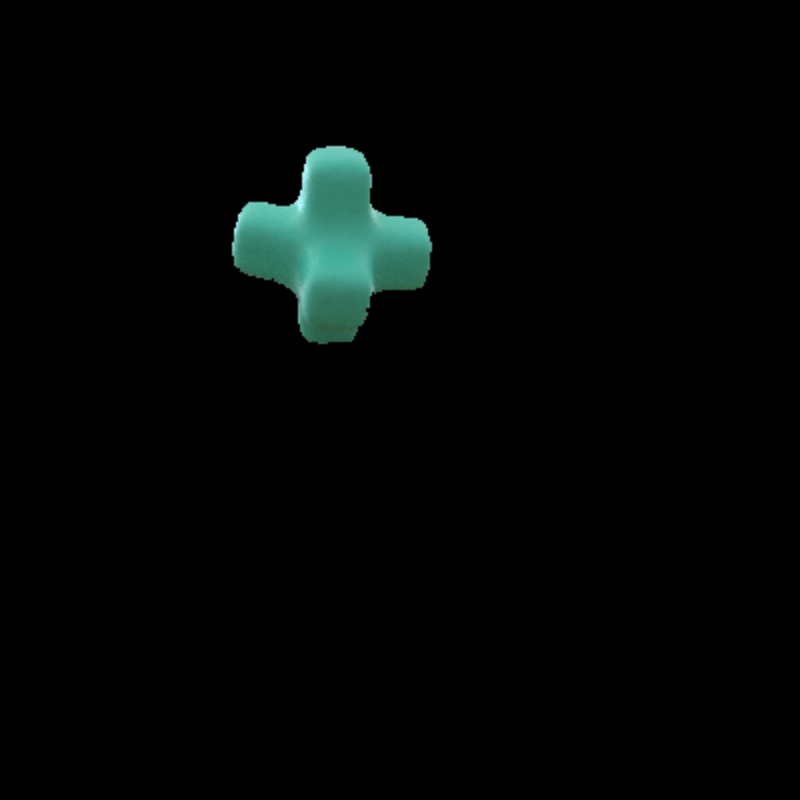}\hspace{-0.275em}
\end{minipage}

% GIC
\begin{minipage}[c]{0.035\textwidth}
  \rotatebox{90}{\textbf{GIC}~\cite{cai2024gic}}
\end{minipage}%
\hspace{-1em}%  
\begin{minipage}[c]{0.49\textwidth}
  \includegraphics[width=0.24\textwidth]{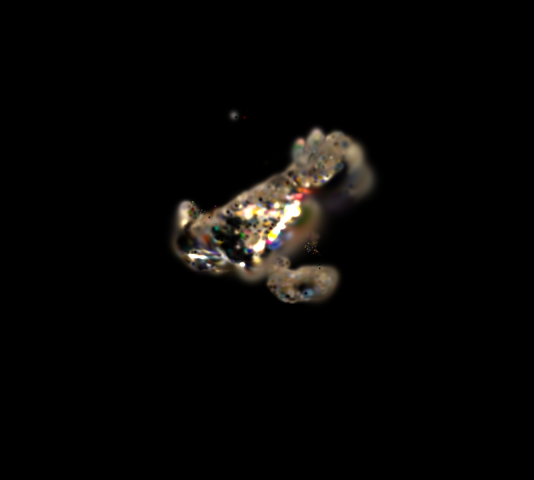}%
  \includegraphics[width=0.24\textwidth]{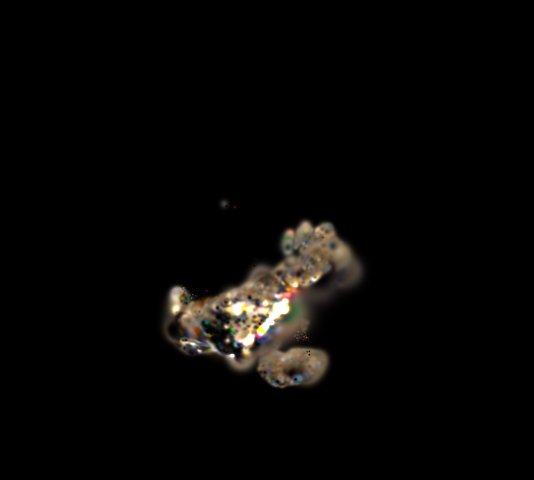}%
  \includegraphics[width=0.24\textwidth]{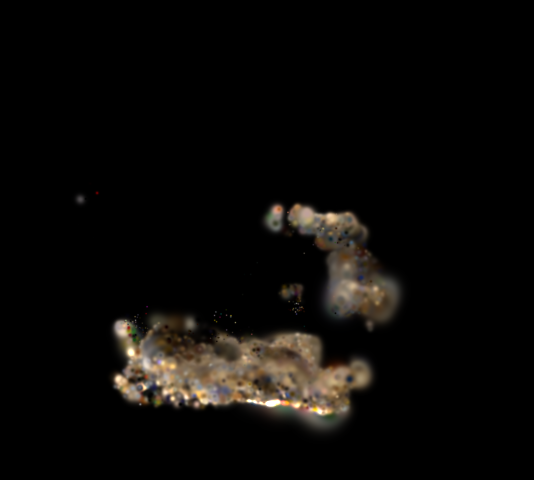}%
  \includegraphics[width=0.24\textwidth]{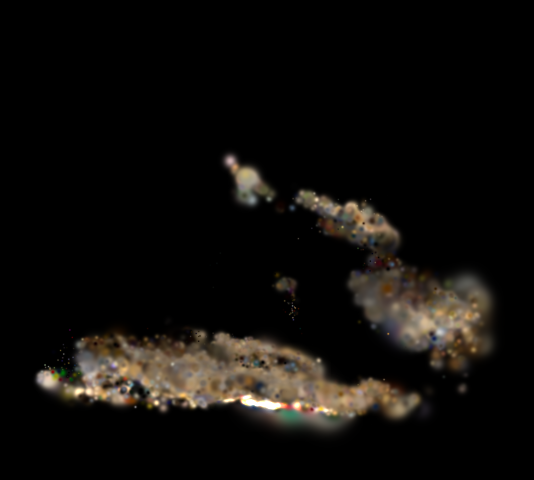}%
\end{minipage}%
% \hspace{2mm}%
\begin{minipage}[c]{0.49\textwidth}
  \includegraphics[trim=130 300 130 0, clip, width=0.24\textwidth]{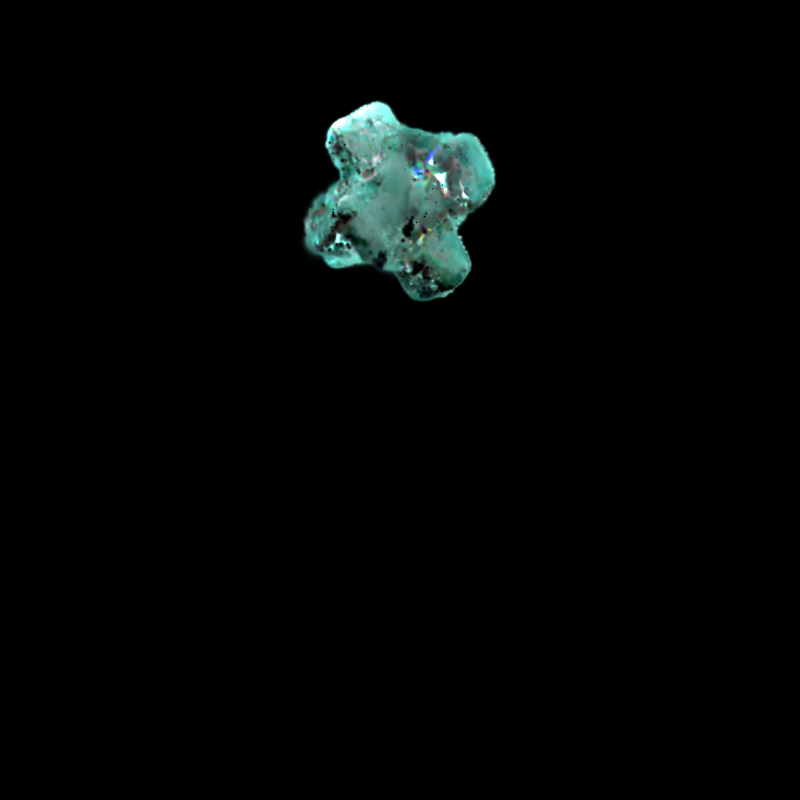}\hspace{-0.275em}
  \includegraphics[trim=130 300 130 0, clip, width=0.24\textwidth]{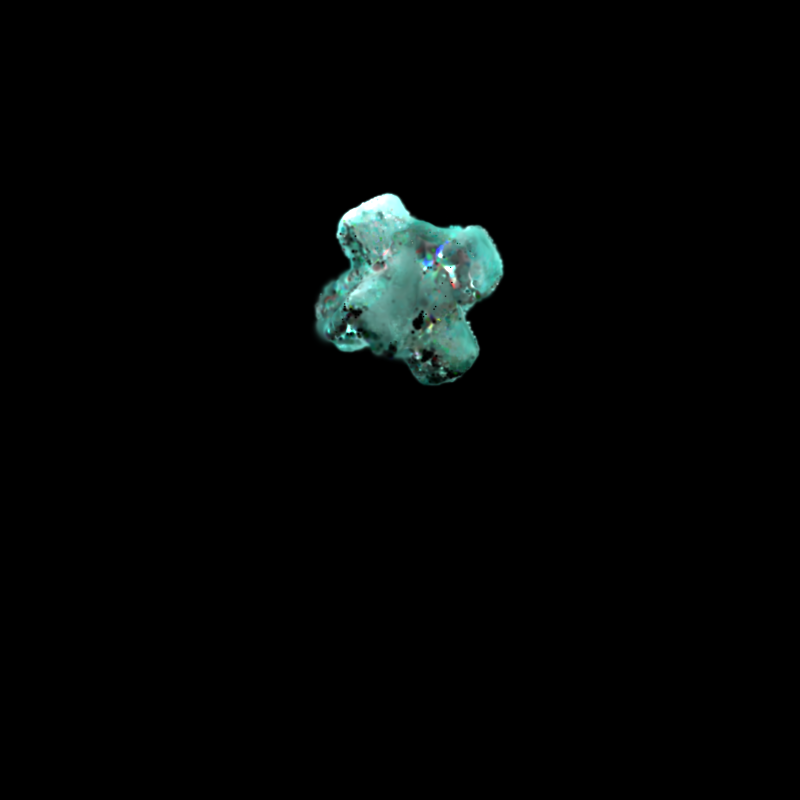}\hspace{-0.275em}
  \includegraphics[trim=130 300 130 0, clip, width=0.24\textwidth]{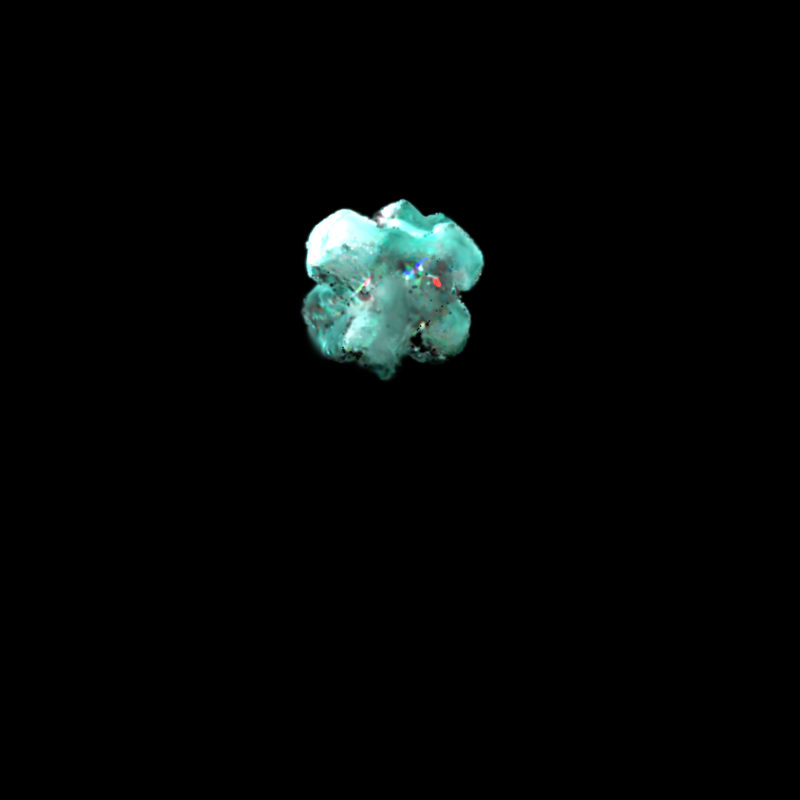}\hspace{-0.275em}
  \includegraphics[trim=130 300 130 0, clip, width=0.24\textwidth]{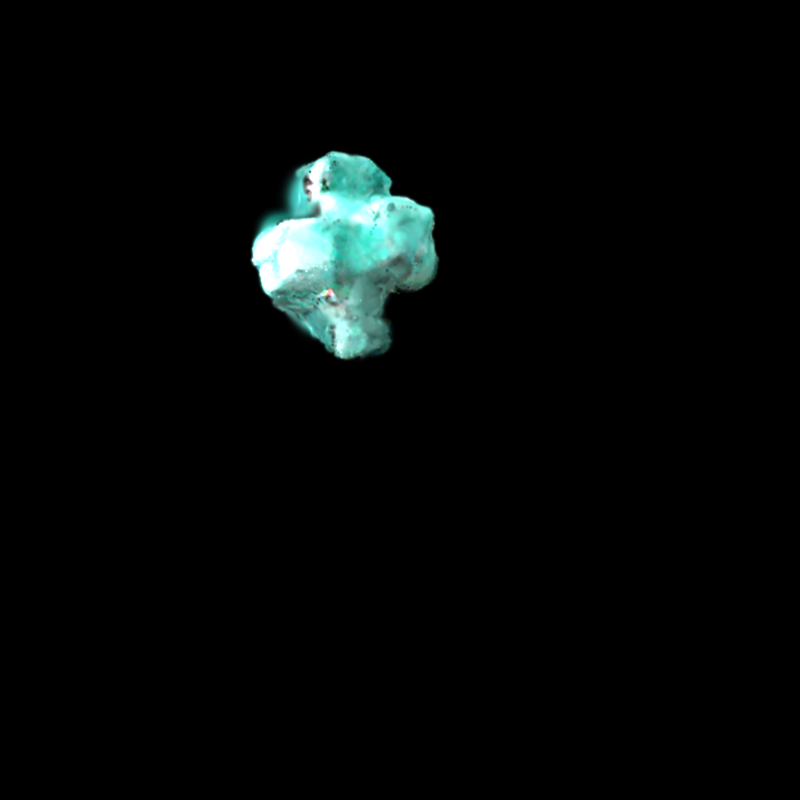}\hspace{-0.275em}
\end{minipage}

% Ours
\begin{minipage}[c]{0.035\textwidth}
  \rotatebox{90}{\textbf{Ours}}
\end{minipage}%
\hspace{-1em}%  
\begin{minipage}[c]{0.49\textwidth}
  % LEFT bottom right top
  \includegraphics[trim=130 300 130 0, clip, width=0.24\textwidth]{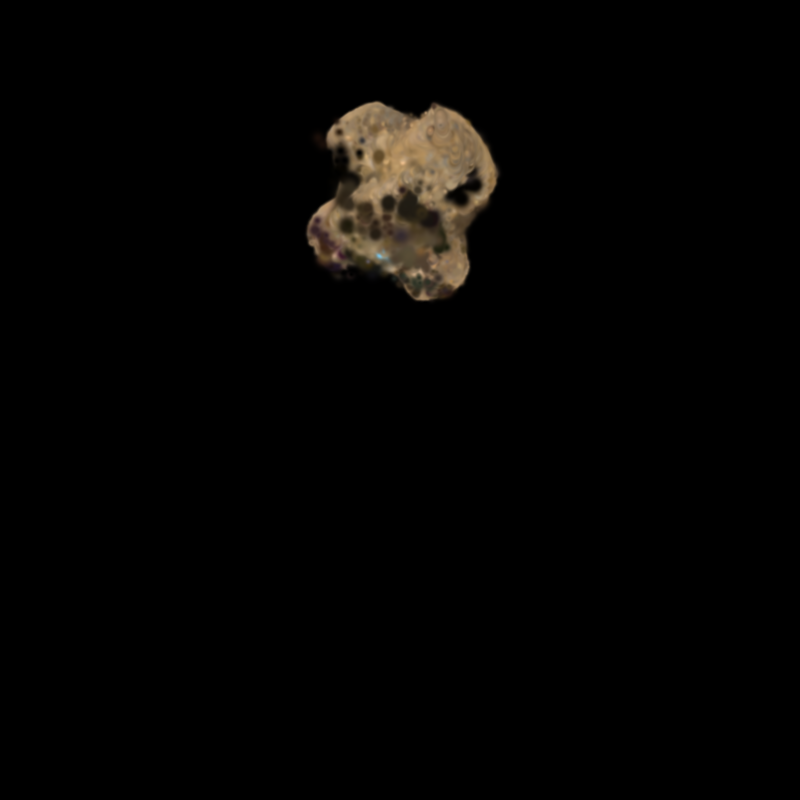}%
  \includegraphics[trim=130 300 130 0, clip, width=0.24\textwidth]{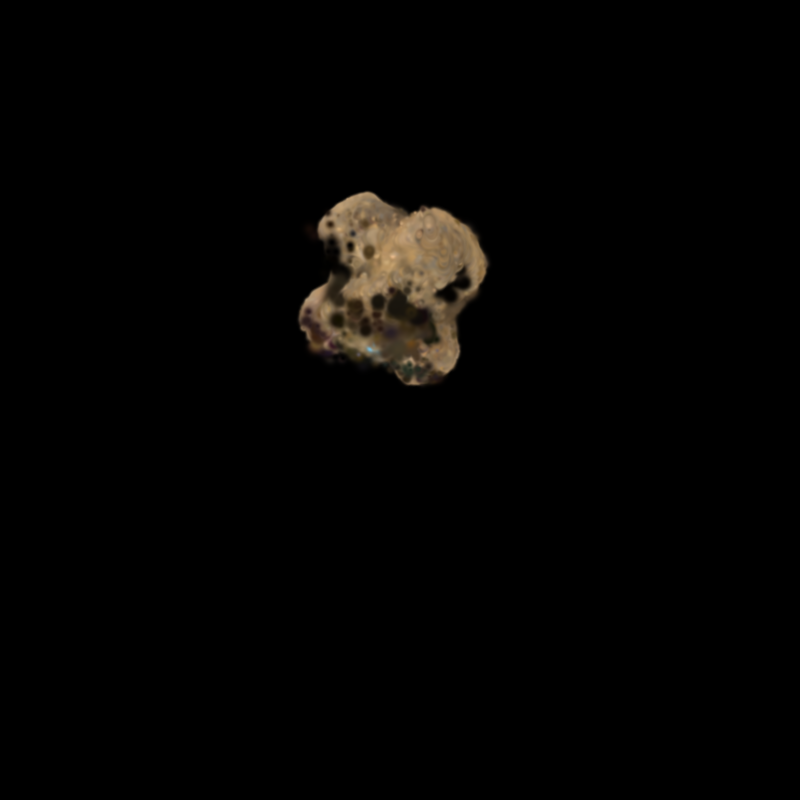}%
  \includegraphics[trim=130 300 130 0, clip, width=0.24\textwidth]{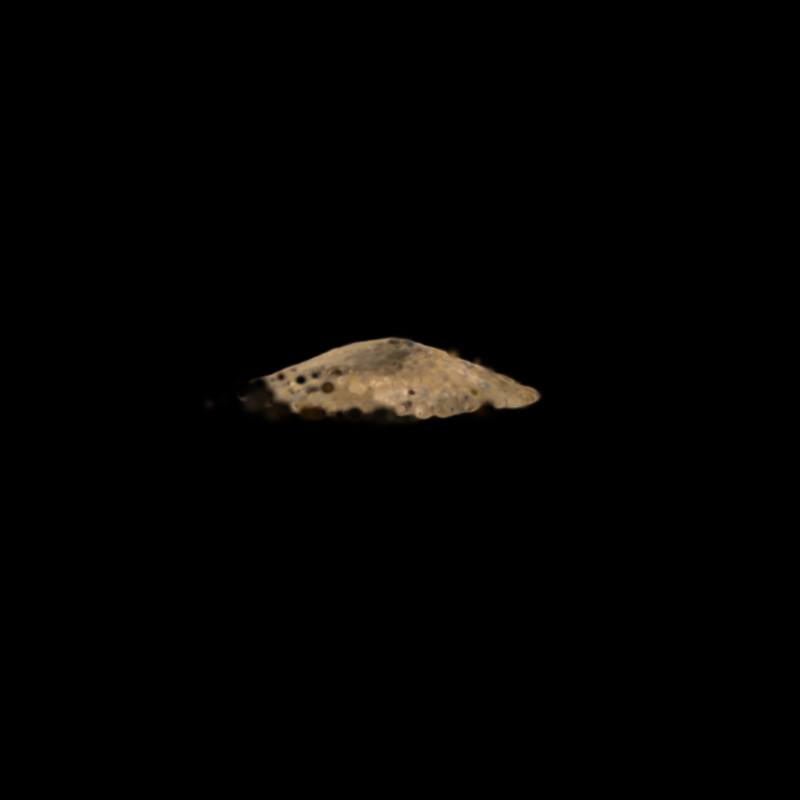}%
  \includegraphics[trim=130 300 130 0, clip, width=0.24\textwidth]{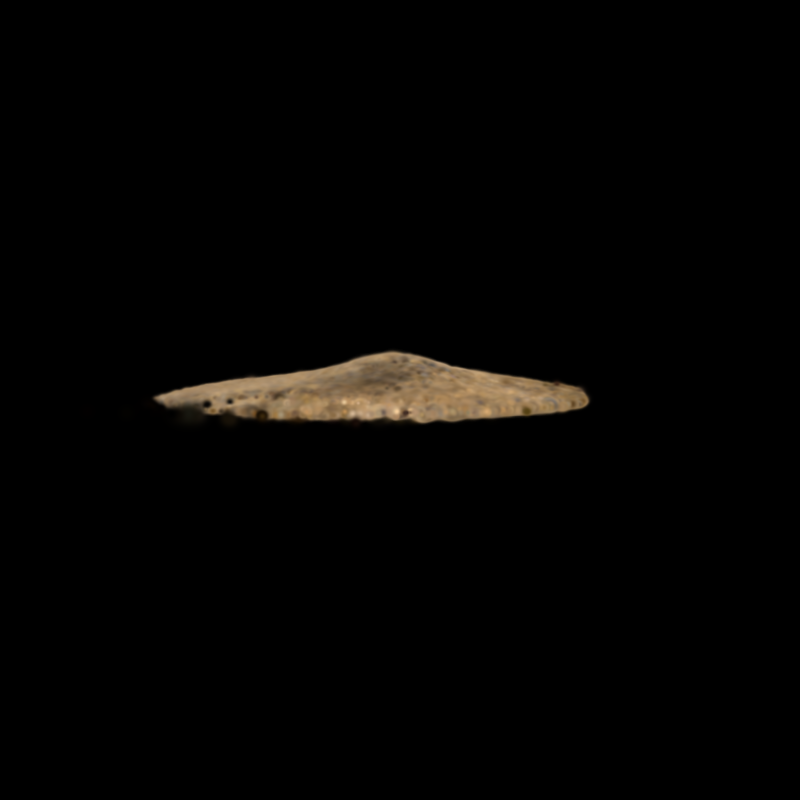}%
\end{minipage}%
% \hspace{2mm}%
\begin{minipage}[c]{0.49\textwidth}
  \includegraphics[trim=130 300 130 0, clip, width=0.24\textwidth]{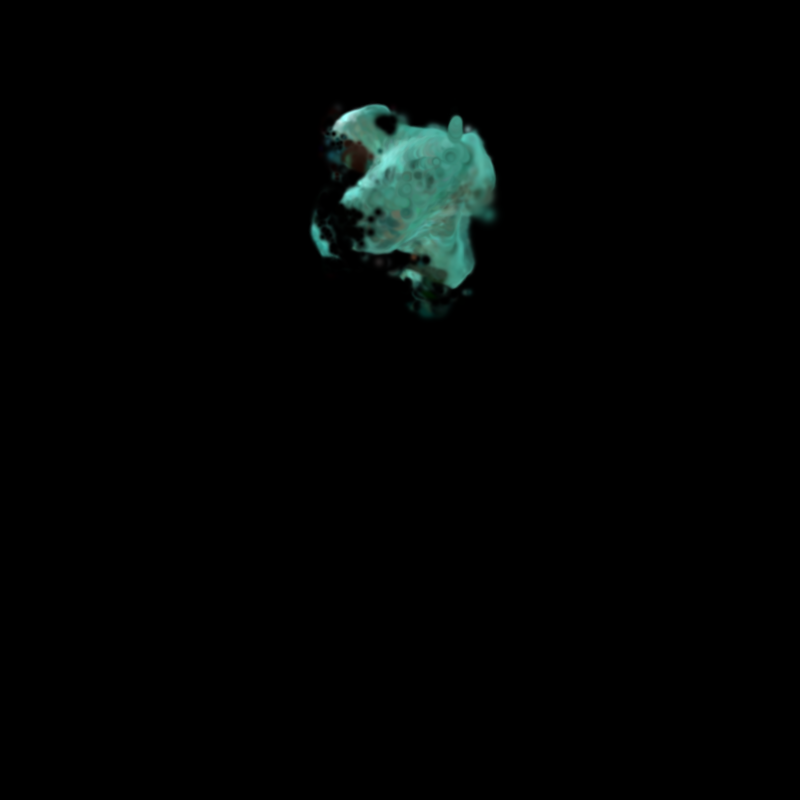}\hspace{-0.275em}
  \includegraphics[trim=130 300 130 0, clip, width=0.24\textwidth]{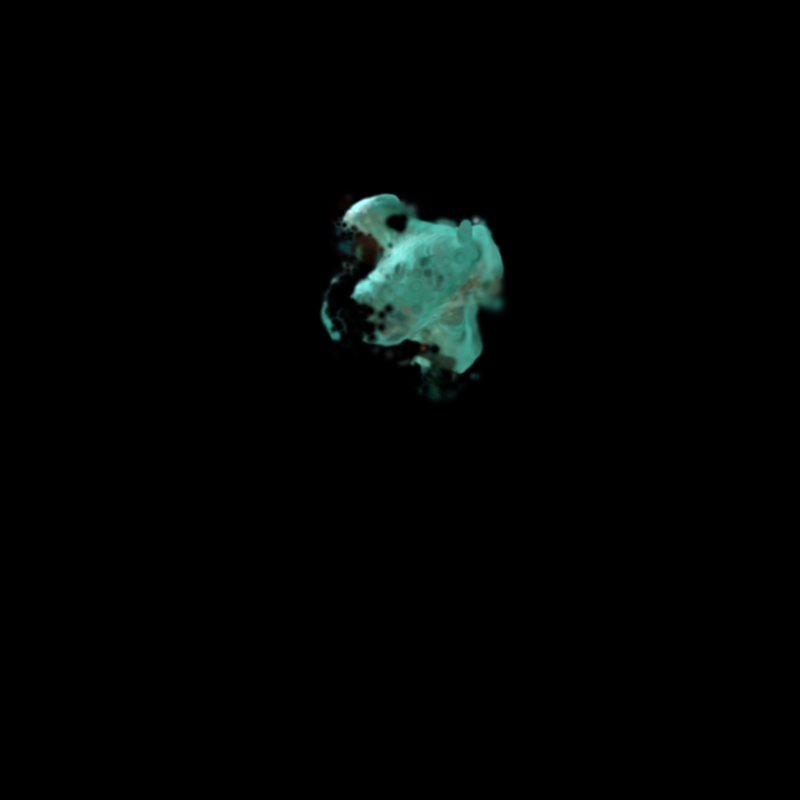}\hspace{-0.275em}
  \includegraphics[trim=130 300 130 0, clip, width=0.24\textwidth]{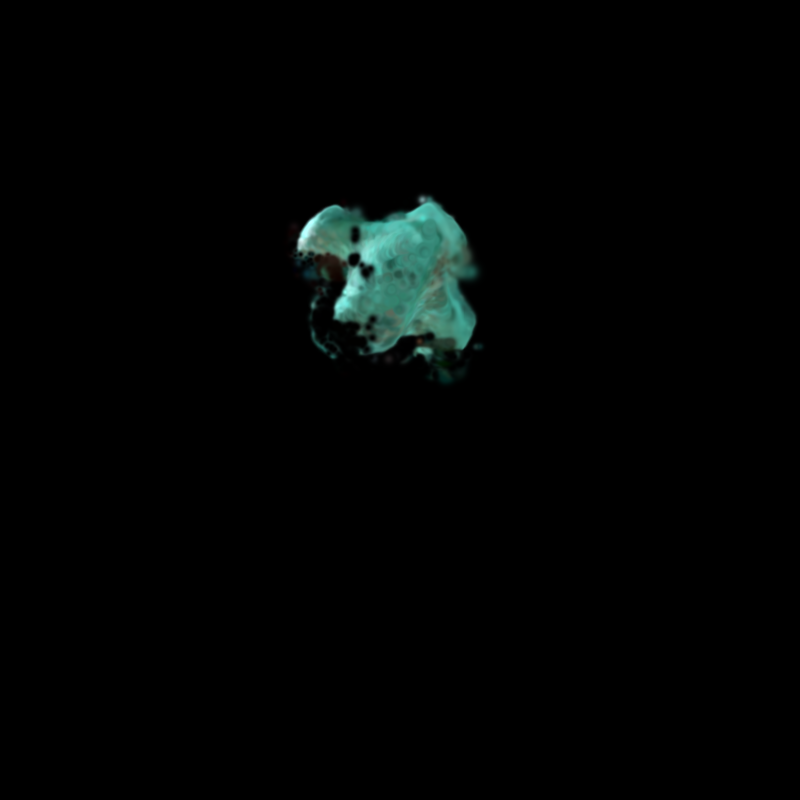}\hspace{-0.275em}
  \includegraphics[trim=130 300 130 0, clip, width=0.24\textwidth]{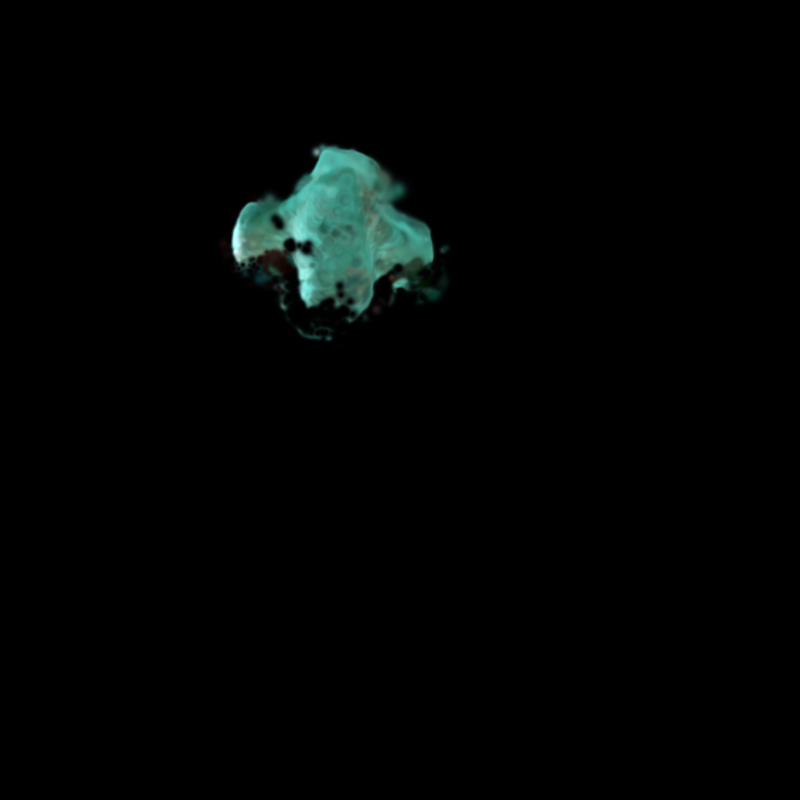}\hspace{-0.275em}
\end{minipage}

\begin{minipage}[c]{0.49\textwidth}
\hspace{9em}%  
  {\textbf{Novel-view Rendering}}
\end{minipage}
\begin{minipage}[c]{0.48\textwidth}
\hspace{9em}%  
  {\textbf{Novel-view Rendering}}
\end{minipage}

\caption{Visual comparison of ProJo4D (Ours) with GIC~\cite{cai2024gic} for novel-view rendering on `Sand' (left) and `elastic' (right) materials from the PAC-NeRF dataset, using sparse-view inputs. ProJo4D reconstructs significantly better geometry than GIC.} % ; Rows 4-6: newtonian (left) and plasticine (right).}
\label{fig:pacnerf}

\end{figure}

\begin{table*}[ht!]
    \caption{Evaluation on PAC-NeRF dataset. We report future prediction accuracy using Chamfer Distance (CD) and Earth Mover's Distance (EMD), and material estimation performance with mean absolute error.}
    \label{tab:pacnerf}
    \centering
    \resizebox{0.95\linewidth}{!}{
        \begin{tabular}{rlrrrrr}
            \toprule
             & & \textbf{Elasticity} & \textbf{Newtonian} & \textbf{Non-Newtonian} & \textbf{Plasticine} & \textbf{Sand} \\
            \midrule
            % ------------------------- CD -------------------------
            \multirow{4}{*}{CD \(\downarrow\)}
            & \revised{Spring-Gaus} & \revised{64.076 $\pm$ 107.271} & - & - & - & - \\
            & GIC & 5.512 $\pm$ 3.311 & 0.537 $\pm$ 0.315 & 0.689 $\pm$ 0.398 & 2.012 $\pm$ 1.797 & 20.262 $\pm$ 43.360 \\
            & ProJo4D & \textbf{0.913 $\pm$ 0.301} & \textbf{0.339 $\pm$ 0.108} & \textbf{0.473 $\pm$ 0.248} & \textbf{1.103 $\pm$ 0.948} & \textbf{0.264 $\pm$ 0.017} \\
            & Full Joint & 1.318 $\pm$ 1.117 & 0.346 $\pm$ 0.095 & 8.104 $\pm$ 13.563 & 17.678 $\pm$ 18.170 & 53.564 $\pm$ 19.404 \\
            \midrule
            % ------------------------- EMD -------------------------
            \multirow{4}{*}{EMD \(\downarrow\)}
            & \revised{Spring-Gaus} & \revised{0.267 $\pm$ 0.154} & - & - & - & - \\
            & GIC & 0.126 $\pm$ 0.041 & 0.103 $\pm$ 0.007 & 0.040 $\pm$ 0.007 & 0.062 $\pm$ 0.027 & 0.122 $\pm$ 0.162 \\
            & ProJo4D & \textbf{0.042 $\pm$ 0.007} & \textbf{0.039 $\pm$ 0.004} & \textbf{0.038 $\pm$ 0.005} & \textbf{0.053 $\pm$ 0.018} & \textbf{0.045 $\pm$ 0.006} \\
            & Full Joint & 0.049 $\pm$ 0.023 & 0.040 $\pm$ 0.005 & 0.099 $\pm$ 0.073 & 0.124 $\pm$ 0.074 & 0.223 $\pm$ 0.020 \\
            \midrule
            % ------------------------- MAE of v -------------------------
            \multirow{4}{*}{MAE $v_0$ \(\downarrow\)}
            & \revised{Spring-Gaus} & \revised{0.292 $\pm$ 0.057} & - & - & - & - \\
            & GIC & 0.008 $\pm$ 0.004 & 0.009 $\pm$ 0.004 & 0.015 $\pm$ 0.008 & \textbf{0.010 $\pm$ 0.005} & 0.007 $\pm$ 0.004 \\
            & ProJo4D & \textbf{0.007 $\pm$ 0.003} & \textbf{0.008 $\pm$ 0.002} & \textbf{0.005 $\pm$ 0.003} & 0.024 $\pm$ 0.056 & \textbf{0.005 $\pm$ 0.003} \\
            & Full Joint & 0.020 $\pm$ 0.033 & 0.008 $\pm$ 0.004 & 0.080 $\pm$ 0.099 & 0.092 $\pm$ 0.102 & 0.046 $\pm$ 0.032 \\
            \midrule
            % ------------------------- MAE of E -------------------------
            \multirow{4}{*}{MAE $\log (E)$ \(\downarrow\)}
            & \revised{Spring-Gaus} & - & - & - & - & - \\
            & GIC & 0.189 $\pm$ 0.217 & -  & -  & 1.597 $\pm$ 1.150 & -  \\
            & ProJo4D & \textbf{0.124 $\pm$ 0.099} & -  & - & \textbf{0.742 $\pm$ 0.780} & -  \\
            & Full Joint & 0.216 $\pm$ 0.299 & - & - & 2.856 $\pm$ 2.196 & - \\
            \midrule
            % ------------------------- MAE of $\nu$ -------------------------
            \multirow{4}{*}{MAE $\nu$ \(\downarrow\)}
            & \revised{Spring-Gaus} & - & - & - & - & - \\
            & GIC & 0.123 $\pm$ 0.103 & -  & -  & 0.134 $\pm$ 0.112 & -  \\
            & ProJo4D & \textbf{0.048 $\pm$ 0.034} & -  & - & \textbf{0.084 $\pm$ 0.029} & -  \\
            & Full Joint & 0.061 $\pm$ 0.053 & - & - & 0.075 $\pm$ 0.057 & - \\
            \midrule
            % ------------------------- MAE of mu -------------------------
            \multirow{4}{*}{MAE $\log (\mu)$ \(\downarrow\)}
            & \revised{Spring-Gaus} & - & - & - & - & - \\
            & GIC & - & \textbf{0.103 $\pm$ 0.125} & 0.869 $\pm$ 0.598 & -& - \\
            & ProJo4D & - & 0.134 $\pm$ 0.175 & \textbf{0.491 $\pm$ 0.363} & - & - \\
            & Full Joint & - & 0.294 $\pm$ 0.314 & 2.315 $\pm$ 1.100 & - & - \\
            \midrule
            % ------------------------- MAE of kappa -------------------------
            \multirow{4}{*}{MAE $\log (\kappa)$ \(\downarrow\)}
            & \revised{Spring-Gaus} & - & - & - & - & - \\
            & GIC & - & 3.180 $\pm$ 1.085 & 0.725 $\pm$ 0.704 & - & - \\
            & ProJo4D & - & \textbf{1.425 $\pm$ 1.148} & \textbf{0.462 $\pm$ 0.344} & - & - \\
            & Full Joint & - & 3.312 $\pm$ 1.679 & 1.673 $\pm$ 1.918 & - & - \\
            \midrule
            % ------------------------- MAE of T_y -------------------------
            \multirow{4}{*}{MAE $\log (\tau_Y)$ \(\downarrow\)}
            & \revised{Spring-Gaus} & - & - & - & - & - \\
            & GIC & - & - & \textbf{0.069 $\pm$ 0.069} & 0.327 $\pm$ 0.365 & - \\
            & ProJo4D & - & - & 0.144 $\pm$ 0.071 & \textbf{0.144 $\pm$ 0.125} & - \\
            & Full Joint & - & - & 1.839 $\pm$ 3.242 & 6.612 $\pm$ 7.739 & - \\
            \midrule
            % ------------------------- MAE of eta -------------------------
            \multirow{4}{*}{MAE $\log (\eta)$ \(\downarrow\)}
            & \revised{Spring-Gaus} & - & - & - & - & - \\
            & GIC & -  & -  & 0.519 $\pm$ 0.264 & -  & -  \\
            & ProJo4D & -  & -  & \textbf{0.463 $\pm$ 0.244} & -  & -  \\
            & Full Joint & - & - & 1.455 $\pm$ 2.263 & - & - \\
            \midrule
            % ------------------------- MAE of friction alpha -------------------------
            \multirow{4}{*}{MAE $\theta_{fric}$ \(\downarrow\)}
            & \revised{Spring-Gaus} & - & - & - & - & - \\
            & GIC & -  & -  & -  & -  & 6.785 $\pm$ 8.458 \\
            & ProJo4D & -  & -  & -  & -  & \textbf{4.998 $\pm$ 2.542} \\
            & Full Joint & - & - & - & - & 67.893 $\pm$ 12.416 \\
            \bottomrule
        \end{tabular}
    }
\end{table*}

\section{Experiments}
\label{sec:exp}

\subsection{Experimental Settings}
\label{ssec:exp_settings}

\textbf{Baselines.}
We compare ours with the current state-of-the-art methods; PAC-NeRF~\cite{li2023pacnerf}, Spring-Gaus~\cite{zhong2024springgaus}, GIC~\cite{cai2024gic}, and Vid2Sim~\cite{Vid2Sim}.

\textbf{Datasets.}
% licenses are also for checklist
We used the synthetic dataset from Spring-Gaus~\cite{zhong2024springgaus}, PAC-NeRF~\cite{li2023pacnerf}, and
the GSO dataset from Vid2Sim~\cite{Vid2Sim}.
In the Spring-Gaus dataset, 7 distinct object shapes are provided, whereas the GSO dataset offers 12 shapes; both datasets only involve elastic materials with varying parameters.
We used the PAC-NeRF synthetic dataset that comprises five different material models: elastic (Neo-Hookean), Newtonian fluid, non-Newtonian fluid, plasticine, and sand (Drucker-Prager), each sharing the same object shape but with different physical parameters.
The dataset has a total of 45 scenes, and we report the mean and standard deviation for each material model.
To evaluate how our proposed optimization strategy can improve performance in sparse-view settings, we selected only three cameras from the ten cameras available for both datasets: the second, sixth, and tenth cameras from each scene using Spring-Gaus and PAC-NeRF datasets.
For the GSO dataset, we used the same experimental settings as Vid2Sim.
To compare with other baselines, we used the same train/test splits provided by each dataset.

% real dataset
For real-world evaluation, we use the Spring-Gaus dataset, which provides exactly three cameras for each scene.
Following Spring-Gaus~\cite{zhong2024springgaus}, we optimize with the same data preprocessing steps.
We train each model on the first 14 frames and test on the remaining 6 frames.

\textbf{Metrics.}
To evaluate our method against state-of-the-art approaches, we adopt two categories of metrics: future state prediction and physical parameter estimation, following prior works~\cite{li2023pacnerf,zhong2024springgaus,cai2024gic}.  
For future state prediction, we measure the 3D discrepancy between simulated positions $\tilde{\mathbf{x}}_t$ and ground-truth positions $x_t$ using Chamfer Distance (CD) and Earth Mover's Distance (EMD). We also assess the 2D rendering quality of predicted future states from both seen viewpoints (three in our experiments) and novel viewpoints using peak signal-to-noise ratio (PSNR) and structural similarity index measure (SSIM).  
For physical parameter estimation, we report mean absolute error (MAE), following previous literature~\cite{li2023pacnerf,zhong2024springgaus,cai2024gic}.

\textbf{Hyperparameters.}
To focus on the optimization strategy, we use the same learning rates as GIC for both Spring-Gaus and the PAC-NeRF datasets.
We also optimized with the same number of iterations for Stage 1 and 2 as GIC.
We set the loss weights $\lambda_{img}$ and $\lambda_{geo}$ to $1 / |\mathcal{C}|$ and 1.0, respectively (Eqs.~\ref{eq:stage1} and \ref{eq:stage2}).

\subsection{Results}
\label{ssec:results}

\textbf{Synthetic Datasets.}
We evaluate ProJo4D on three synthetic datasets covering diverse object shapes, appearances, and material models.

The Spring-Gaus dataset evaluates performance across diverse object shapes and appearances.
As shown in Tab.~\ref{tab:springgaus}, both Spring-Gaus and GIC degrade significantly under sparse views, especially on trajectory-sensitive scenes such as paste, whereas our method remains considerably more robust.
Fig.~\ref{fig:springgaus} further illustrates that existing approaches fail to estimate geometry and physical parameters, critically failing to predict future trajectories.
In contrast, our method remains robust with sparse views, achieving a tenfold reduction in Chamfer Distance, roughly half the Earth Mover's Distance on average, and a substantial PSNR boost (17.58 $\rightarrow$ 22.30).
% MASIV achieves strong performance (CD: 2.77, EMD: 0.082) by introducing neural material models and additional loss function, but this incurs increased GPU memory requirements and longer optimization times.
MASIV achieves strong performance (CD: 2.77, EMD: 0.082) by introducing neural material models and additional loss function, but \revised{this requires around a day of training per elastic object compared to an hour for ProJo4D on an NVIDIA A6000}.
ProJo4D achieves better accuracy (CD: 1.60, EMD: 0.057) using the same scene representation and physics framework as GIC, with performance improvements attributable solely to our progressive joint optimization strategy.
This demonstrates that careful optimization design can outperform architectural innovations while maintaining computational efficiency and physical interpretability.

The PAC-NeRF dataset evaluates the accuracy and robustness to different material models and their parameters.
Tab.~\ref{tab:pacnerf} and Fig.~\ref{fig:pacnerf} show both quantitatively and qualitatively that our method outperforms GIC across different material models and different metrics and parameters.
Both non-Newtonian and Plasticine have non-differentiable branches, which pose additional challenges to the progressive joint optimization strategy.
Nevertheless, our method improves upon GIC in 4/5 parameters for Non-Newtonian and in 3/4 parameters for Plasticine, and shows strong improvement in 3D reconstruction (CD \& EMD).

\begin{table}[t]
    \centering
    \begin{minipage}{0.43\linewidth}
        \centering
        \caption{Future state prediction and material parameter estimation on the GSO dataset.}
        \label{tab:gso}
        \footnotesize
        \setlength{\tabcolsep}{4pt}
        \resizebox{\linewidth}{!}{
            \begin{tabular}{lccc}
                \toprule
                 & PSNR \(\uparrow\) & MAE $\log E$ \(\downarrow\) & MAE $\nu$ \(\downarrow\) \\
                \midrule
                PAC-NeRF & 20.11 & 2.50 & 0.21 \\
                Spring-Gaus & 18.32 & - & - \\
                GIC & 19.20 & 2.01 & 0.16 \\
                Vid2Sim & 25.07 & 0.51 & \textbf{0.06} \\
                \cline{1-4}
                GIC* & 21.90 & 0.59 & 0.07 \\
                GIC* + ProJo4D & \textbf{26.80} & \textbf{0.31} & \textbf{0.06} \\
                \bottomrule
            \end{tabular}
        }
    \end{minipage}%
    \hfill
    \begin{minipage}{0.55\linewidth}
        \centering
        \caption{2D future state prediction accuracies on Spring-Gaus real-world dataset~\cite{zhong2024springgaus}.}
        \label{tab:real}
        \footnotesize
        \setlength{\tabcolsep}{3pt}
        \resizebox{\linewidth}{!}{
            \begin{tabular}{clrrrrr r}
                \toprule
                 & & \textbf{bun} & \textbf{burger} & \textbf{dog} & \textbf{pig} & \textbf{potato} & \textbf{mean} \\
                \midrule
                \multirow{3}{*}{PSNR \(\uparrow\)}
                & Spring-Gaus & 26.79 & 35.13 & 30.31 & 31.95 & 28.96 & 30.63 \\
                & GIC & 32.14 & 36.89 & 33.35 & 32.30 & 35.02 & 34.05 \\
                & GIC + ProJo4D & \textbf{37.35} & \textbf{39.01} & \textbf{36.07} & \textbf{38.90} & \textbf{40.18} & \textbf{38.30} \\
                \midrule
                \multirow{3}{*}{SSIM \(\uparrow\)}
                & Spring-Gaus & 0.986 & 0.995 & 0.993 & 0.994 & 0.989 & 0.991 \\
                & GIC & 0.994 & 0.995 & 0.995 & 0.996 & 0.995 & 0.995 \\
                & GIC + ProJo4D & \textbf{0.997} & \textbf{0.996} & \textbf{0.996} & \textbf{0.997} & \textbf{0.997} & \textbf{0.996} \\
                \bottomrule
            \end{tabular}
        }
    \end{minipage}
\end{table}

We also evaluate our method on the GSO dataset from Vid2Sim~\cite{Vid2Sim}.
To ensure a fair comparison, we reran GIC and ProJo4D using the same material model as Vid2Sim on the GSO dataset.
Although Vid2Sim, PAC-NeRF, and GIC all adopt the Neo-Hookean material model for elastic objects, their exact formulations differ.
The variant, GIC*, in Tab.~\ref{tab:gso} reports the performance of GIC with the identical material model as Vid2Sim.
Results indicate that aligning the material models leads to improved performance for GIC.
Our proposed method, ProJo4D, consistently demonstrates further performance improvements.

\revised{In summary, ProJo4D broadly improves over existing methods across different datasets, material models, and object shapes, with strong improvement in future trajectory prediction.}
In multi-parameter inverse problems, multiple parameter configurations can produce nearly indistinguishable dynamics in simulation, as physical parameters are often coupled.
The future state prediction accuracy, which reflects the combined effect of all estimated parameters, is a more indicative measure of overall estimation quality.
Examples are ``banana'' and ``paste'' in the Spring-Gaus dataset: while ProJo4D underperforms GIC in a single physics parameter, it achieves lower CD and EMD and higher PSNR in future prediction.
ProJo4D not only improves parameter estimation on average but, more importantly, delivers better future state prediction accuracy, which is especially relevant for downstream applications such as simulation and digital twin construction.
% In summary, ProJo4D shows consistent improvement over state-of-the-art methods across different datasets, material models and object shapes.
% In multi-parameter inverse problems, exact recovery of every parameter is often ill-posed: multiple parameter configurations can generate nearly indistinguishable dynamics in future state simulation.
% Worse performance in one parameter alongside improvement in others does not necessarily indicate worse overall inverse physics estimation.
% The future state prediction accuracy is often a more reliable indicator of the overall estimation quality.
% Examples are ``banana'' and ``paste'' in Spring-Gaus dataset: while ProJo4D underperforms GIC in a single physics parameter, it consistently achieves lower CD and EMD and higher PSNR in future prediction.
% Taken together, ProJo4D not only improves parameter estimation overall but, more importantly, delivers reliable gains in future state predictive accuracy, an outcome that is especially relevant for downstream applications such as simulation and digital twin construction.

\textbf{Real-world Dataset.}
We evaluate ProJo4D on the real-world Spring-Gaus dataset to demonstrate practical applicability.
Tab.~\ref{tab:real} provides results on the Spring-Gaus real-world dataset.
Since no ground truth three-dimensional mesh or material parameters are available, we evaluate using only two-dimensional metrics: PSNR and SSIM.
Because Spring-Gaus works exclusively with elastic objects, we use the elastic material model for both GIC~\cite{cai2024gic} and our method.
Consistent with the synthetic data results in Tabs.~\ref{tab:springgaus} and \ref{tab:pacnerf}, our method outperforms other approaches in both PSNR and SSIM on real-world images.
This demonstrates that our proposed optimization strategy significantly enhances estimation performance not only in synthetic but also in real-world settings, with additional visual results provided in the appendix.

\begin{table}[ht]
    \centering
    \footnotesize
    \setlength{\tabcolsep}{4pt}
    \caption{
        \textbf{Comparison with alternative optimization strategies.} 
        Optimization order is denoted by X (position), A (appearance), S (velocity), and M (material parameters). 
        ``Sequential'' is the GIC baseline. ``Sequential+'' increases iterations to match ProJo4D's budget. 
        ``Cyclic'' strategies (N=4) repeat optimization stages with fewer iterations per cycle.
    }
    \label{tab:abl_alts}
    \begin{tabular}{rr|rr|rr}
        \toprule
         & ProJo4D  & Sequential & Sequential+ & \multicolumn{2}{c}{Cyclic} \\
         & XA-XAS-XASM & XA-S-M-A & XA-S-M-A & (XA-S-M)$\times$4-A & XA-(S-M-A)$\times$4 \\
        \midrule
        CD \(\downarrow\) & \textbf{1.60} & 16.11 & 16.72 & 14.63 & 3.20 \\
        EMD \(\downarrow\) & \textbf{0.057} & 0.128 & 0.135 &  0.136 & 0.085 \\
        \midrule
        PSNR \(\uparrow\) & \textbf{22.30} & 17.58 & 18.01 &  16.96 & 19.04 \\
        SSIM \(\uparrow\) & \textbf{0.913} & 0.850 & 0.854 &  0.834 & 0.882 \\
        \midrule
        MAE $\log E$ \(\downarrow\) & \textbf{0.1043} & 0.2311 & 0.1958 & 0.2547 & 0.1616 \\
        MAE $\nu$ \(\downarrow\) & \textbf{0.0911} & 0.1790 & 0.2984 & 0.2524 & 0.2686 \\
        \bottomrule
    \end{tabular}

\end{table}

\begin{table}[t]
    \centering
    \footnotesize
    \setlength{\tabcolsep}{4pt}
    \caption{
        Ablation study on the impact of different numbers of camera views.
        % ProJo4D consistently improves performance across varying numbers of views.
    }
    \label{tab:abl_n_views}
    % \resizebox{\linewidth}{!}{
        \begin{tabular}{rlrrrr}
            \toprule
            & & \multicolumn{4}{c}{Number of cameras} \\
            \cmidrule(lr){3-6}
             & & \revised{1} & 2 & 3 & 10 \\
            \midrule
            \multirow{2}{*}{\revised{Init CD \(\downarrow\)}}
            & \revised{GIC} & \revised{27.08} & \revised{0.60} & \revised{0.40} & \revised{\textbf{0.10}} \\
            & \revised{+ ProJo4D} & \revised{\textbf{26.79}} & \revised{\textbf{0.52}} & \revised{\textbf{0.34}} & \revised{0.11} \\
            \midrule
            \multirow{2}{*}{Test CD \(\downarrow\)}
            & GIC & \revised{107.20} & 12.00 & 16.11 & 0.95 \\
            & + ProJo4D & \revised{\textbf{57.74}} & \textbf{1.66} & \textbf{1.60} & \textbf{0.65} \\
            % \midrule
            % \multirow{2}{*}{EMD \(\downarrow\)}
            % & GIC & 0.129 & 0.128 & 0.049 \\
            % & + ProJo4D & \textbf{0.055} & \textbf{0.057} & \textbf{0.034} \\
            \midrule
            \multirow{2}{*}{PSNR \(\uparrow\)}
            & GIC & \revised{11.26} & 16.57 & 17.58 & 22.98 \\
            & + ProJo4D & \revised{\textbf{15.48}} & \textbf{20.56} & \textbf{22.30} & \textbf{26.95} \\
            % \midrule
            % \multirow{2}{*}{SSIM \(\uparrow\)}
            % & GIC & 0.844 & 0.850 & 0.930\\
            % & + ProJo4D & \textbf{0.880} & \textbf{0.913} & \textbf{0.951} \\
            \midrule
            \multirow{2}{*}{MAE $\log E$ \(\downarrow\)}
            & GIC & \revised{0.5807} & 0.4951 & 0.2311 & 0.1286 \\
            & + ProJo4D & \revised{\textbf{0.2527}} & \textbf{0.1094} & \textbf{0.1043} & \textbf{0.0643} \\
            \midrule
            \multirow{2}{*}{MAE $\nu$ \(\downarrow\)}
            & GIC & \revised{0.2194} & 0.2752 & 0.1790 & \textbf{0.0458} \\
            & + ProJo4D & \revised{\textbf{0.1434}} & \textbf{0.1374} & \textbf{0.0911} & 0.0654 \\
            \bottomrule
        \end{tabular}
    % }
\end{table}

\subsection{Ablation Study}
\label{ssec:ablation}

% We conduct ablation studies to analyze the effectiveness of our progressive joint optimization strategy and evaluate its robustness under different conditions.

\textbf{Optimization Strategy Comparison.}
As motivated in Sec.~\ref{ssec:motivation}, the choice of optimization strategy significantly impacts performance.
Tab.~\ref{tab:pacnerf} shows that full joint optimization fails for complex materials like Non-Newtonian, Plasticine, and Sand, despite working reasonably for simpler materials.
Here, we investigate whether alternative scheduling strategies can match ProJo4D's progressive approach.

Tab.~\ref{tab:abl_alts} compares ProJo4D against sequential and cyclic strategies on the Spring-Gaus dataset.
``Sequential'' refers to the baseline GIC. % , which optimizes parameters in separate stages without joint refinement.
To ensure a fair comparison, we also evaluate ``Sequential+,'' which matches ProJo4D's total iteration count and the number of images used per parameter set (detailed configurations in the appendix, Tab.~\ref{tab:abl_alts_details}).
We further evaluate cyclic optimization, where parameter sets are optimized one at a time over multiple cycles while maintaining a constant total iteration budget:
\begin{enumerate}
    \vspace{-8pt}
    \item \textbf{(XA-S-M)$\times$4-A}: Cycles through 4D learning, velocity, and material optimization 4 times, followed by appearance refinement.
    \vspace{-8pt}
    \item \textbf{XA-(S-M-A)$\times$4}: Performs 4D representation learning first, then cycles through velocity, material, and appearance optimization 4 times.
    \vspace{-8pt}
\end{enumerate}
The results show that simply increasing the number of iterations (Sequential+) does not improve meaningfully, confirming that the improvement of ProJo4D stems from the optimization strategy.
Cyclic optimization can outperform sequential strategies when applied after 4D representation learning (XA-(S-M-A)$\times$4), but introducing it too early ((XA-S-M)$\times$4-A) degrades performance, as the 4D representation is not yet sufficiently accurate.
Nevertheless, cyclic optimization remains less stable and less accurate than ProJo4D's progressive joint optimization.
Additional ablations on Stage 1 design choices are provided in the appendix.

\textbf{Robustness to Camera Views.}
We additionally evaluate robustness with respect to the number of camera views in Tab.~\ref{tab:abl_n_views}.
As shown in the table, future prediction and material parameter estimation performance improve as the number of camera views increases.
By adjusting only the parameter sets optimized at each stage, the performance remains robust even when the number of views decreases.
\revised{We also report Init CD, the Chamfer Distance at the initial frame. Init CD shows that the error accumulation from geometry inherent in sequential optimization can be alleviated by our proposed optimization strategy.}

\textbf{Generalization to Other Frameworks.}
To demonstrate the applicability of ProJo4D to other methods, we evaluate it on the Spring-Gaus backbone, which differs from GIC in several design choices (e.g., anchors vs. Gaussians for simulation). We adapt the ProJo4D strategy by jointly optimizing velocity, positions, and appearance in the first stage, and all physical parameters in the second stage. With these modifications, Chamfer Distance for 4D future prediction improves from 24.54 to 11.89, compared to 12.83 with the same structural modifications but without progressive optimization. The smaller gains compared to GIC are consistent with Spring-Gaus's constraints: fixed anchor connections and reduced 3D spatial flexibility limit the benefits of progressive joint optimization.

\section{Conclusion}
\label{sec:conclusion}

We introduced ProJo4D, a progressive joint optimization framework that incrementally expands the set of jointly optimized parameters. This strategy ensures robust estimation of geometry, appearance, and physical parameters under highly ambiguous, sparse-view inputs.
Evaluations on benchmark datasets show that ProJo4D consistently outperforms state-of-the-art methods in 4D future state prediction, novel view rendering, and physical parameter estimation, demonstrating practical relevance.

While ProJo4D shows strong performance in 4D scene reconstruction and inverse physics estimation from sparse-view videos, it shares limitations common to existing methods.
First, it cannot overcome fundamental challenges from underlying material models, such as non-differentiable, material-parameter-dependent branches, which require longer iterations and increase sensitivity to learning rates for some, including non-Newtonian fluids.
Second, reliance on computationally intensive physics simulations % , specifically MPM,
results in high costs.
Future work should explore accelerating simulations via neural surrogates or other lightweight methods.
Our results highlight that optimization strategy is an important but often overlooked component in inverse physics pipelines, and we hope this work encourages further investigation in this direction.

\subsubsection*{Acknowledgments}
This work is supported by a National Institute of Health (NIH) project \#R21EB035832 "Next-gen 3D Modeling of Endoscopy Videos".

\bibliography{main}

@inproceedings{bengio2009curriculum,
  title={Curriculum learning},
  author={Bengio, Yoshua and Louradour, J{\'e}r{\^o}me and Collobert, Ronan and Weston, Jason},
  booktitle={Proceedings of the 26th annual international conference on machine learning},
  pages={41--48},
  year={2009}
}

@InProceedings{Vid2Sim,
    author    = {Chen, Chuhao and Dou, Zhiyang and Wang, Chen and Huang, Yiming and Chen, Anjun and Feng, Qiao and Gu, Jiatao and Liu, Lingjie},
    title     = {Vid2Sim: Generalizable, Video-based Reconstruction of Appearance, Geometry and Physics for Mesh-free Simulation},
    booktitle = {Proceedings of the IEEE/CVF Conference on Computer Vision and Pattern Recognition (CVPR)},
    month     = {June},
    year      = {2025},
    pages     = {26545-26555}
}

@article{liu2025physflow,
  title={Unleashing the Potential of Multi-modal Foundation Models and Video Diffusion for 4D Dynamic Physical Scene Simulation},
  author={Liu, Zhuoman and Ye, Weicai and Luximon, Yan and Wan, Pengfei and Zhang, Di},
  journal={CVPR},
  year={2025}
}

@InProceedings{Kaneko_2024_CVPR,
    author    = {Kaneko, Takuhiro},
    title     = {Improving Physics-Augmented Continuum Neural Radiance Field-Based Geometry-Agnostic System Identification with Lagrangian Particle Optimization},
    booktitle = {Proceedings of the IEEE/CVF Conference on Computer Vision and Pattern Recognition (CVPR)},
    month     = {June},
    year      = {2024},
    pages     = {5470-5480}
}

@InProceedings{MASIV,
    author    = {Zhao, Yizhou and Chen, Haoyu and Liu, Chunjiang and Li, Zhenyang and Herrmann, Charles and Hur, Junhwa and Li, Yinxiao and Yang, Ming-Hsuan and Raj, Bhiksha and Xu, Min},
    title     = {Toward Material-Agnostic System Identification from Videos},
    booktitle = {Proceedings of the IEEE/CVF International Conference on Computer Vision (ICCV)},
    month     = {October},
    year      = {2025},
    pages     = {5944-5956}
}

@inproceedings{zhang2024physdreamer,
    title={{PhysDreamer}: Physics-Based Interaction with 3D Objects via Video Generation},
    author={Tianyuan Zhang and Hong-Xing Yu and Rundi Wu and Brandon Y. Feng and Changxi Zheng and Noah Snavely and Jiajun Wu and William T. Freeman},
    booktitle={European Conference on Computer Vision},
    year={2024},
    organization={Springer}
}

@InProceedings{physavatar,
    author="Zheng, Yang
    and Zhao, Qingqing
    and Yang, Guandao
    and Yifan, Wang
    and Xiang, Donglai
    and Dubost, Florian
    and Lagun, Dmitry
    and Beeler, Thabo
    and Tombari, Federico
    and Guibas, Leonidas
    and Wetzstein, Gordon",
    editor="Leonardis, Ale{\v{s}}
    and Ricci, Elisa
    and Roth, Stefan
    and Russakovsky, Olga
    and Sattler, Torsten
    and Varol, G{\"u}l",
    title="PhysAvatar: Learning the Physics of Dressed 3D Avatars from Visual Observations",
    booktitle="Computer Vision -- ECCV 2024",
    year="2025",
    publisher="Springer Nature Switzerland",
    address="Cham",
    pages="262--284",
    isbn="978-3-031-72913-3"
}

@article{zhong2024springgaus,
    title     = {Reconstruction and Simulation of Elastic Objects with Spring-Mass 3D Gaussians},
    author    = {Zhong, Licheng and Yu, Hong-Xing and Wu, Jiajun and Li, Yunzhu},
    journal   = {European Conference on Computer Vision (ECCV)},
    year      = {2024}
}

@inproceedings{nerf,
    author = {Mildenhall, Ben and Srinivasan, Pratul P. and Tancik, Matthew and Barron, Jonathan T. and Ramamoorthi, Ravi and Ng, Ren},
    title = {NeRF: Representing Scenes as Neural Radiance Fields for View Synthesis},
    year = {2020},
    isbn = {978-3-030-58451-1},
    publisher = {Springer-Verlag},
    address = {Berlin, Heidelberg},
    url = {https://doi.org/10.1007/978-3-030-58452-8_24},
    doi = {10.1007/978-3-030-58452-8_24},
    booktitle = {Computer Vision – ECCV 2020: 16th European Conference, Glasgow, UK, August 23–28, 2020, Proceedings, Part I},
    pages = {405–421},
    numpages = {17},
    location = {Glasgow, United Kingdom}
}

@inproceedings{
    gao2025seeing,
    title={Seeing the Wind from a Falling Leaf},
    author={Zhiyuan Gao and Jiageng Mao and Hong-Xing Yu and Haozhe Lou and Emily Yue-ting Jia and Jernej Barbic and Jiajun Wu and Yue Wang},
    booktitle={The Thirty-ninth Annual Conference on Neural Information Processing Systems},
    year={2025},
    url={https://openreview.net/forum?id=4NaW9mbTqq}
}

@inproceedings{
    cai2024gic,
    title={{GIC}: Gaussian-Informed Continuum for Physical Property Identification and Simulation},
    author={Junhao Cai and Yuji Yang and Weihao Yuan and Yisheng HE and Zilong Dong and Liefeng Bo and Hui Cheng and Qifeng Chen},
    booktitle={The Thirty-eighth Annual Conference on Neural Information Processing Systems},
    year={2024},
    url={https://openreview.net/forum?id=SSCtCq2MH2}
}

@inproceedings{NEURIPS2024_510cfd99,
 author = {Wang, Jiaxu and Sun, Jingkai and He, Junhao and Zhang, Ziyi and Zhang, Qiang and Sun, Mingyuan and Xu, Renjing},
 booktitle = {Advances in Neural Information Processing Systems},
 editor = {A. Globerson and L. Mackey and D. Belgrave and A. Fan and U. Paquet and J. Tomczak and C. Zhang},
 pages = {45703--45736},
 publisher = {Curran Associates, Inc.},
 title = {DEL: Discrete Element Learner for Learning 3D Particle Dynamics with Neural Rendering},
 url = {https://proceedings.neurips.cc/paper_files/paper/2024/file/510cfd9945f8bde6f0cf9b27ff1f8a76-Paper-Conference.pdf},
 volume = {37},
 year = {2024}
}

@inproceedings{
    li2023nvfi,
    title={{NVF}i: Neural Velocity Fields for 3D Physics Learning from Dynamic Videos},
    author={Jinxi Li and Ziyang Song and Bo Yang},
    booktitle={Thirty-seventh Conference on Neural Information Processing Systems},
    year={2023},
    url={https://openreview.net/forum?id=gsi9lJ3994}
}

@inproceedings{
    xue2023dintphys,
    title={3D-IntPhys: Towards More Generalized 3D-grounded Visual Intuitive Physics under Challenging Scenes},
    author={Haotian Xue and Antonio Torralba and Joshua B. Tenenbaum and Daniel LK Yamins and Yunzhu Li and Hsiao-Yu Tung},
    booktitle={Thirty-seventh Conference on Neural Information Processing Systems},
    year={2023},
    url={https://openreview.net/forum?id=Fp5uC6YHwe}
}

@inproceedings{yu2023hyfluid,
    title={Inferring hybrid neural fluid fields from videos},
    author={Hong-Xing Yu and Yang Zheng and Yuan Gao and Yitong Deng and Bo Zhu and Jiajun Wu},
    booktitle={NeurIPS},
    year={2023}
}

@inproceedings{qiao2022neuphysics,
    author  = {Qiao, Yi-Ling and Gao, Alexander and Lin, Ming C.},
    title  = {NeuPhysics: Editable Neural Geometry and Physics from Monocular Videos},
    booktitle = {Conference on Neural Information Processing Systems (NeurIPS)},
    year  = {2022},
}

@article{zhong2021extending,
  title={Extending lagrangian and hamiltonian neural networks with differentiable contact models},
  author={Zhong, Yaofeng Desmond and Dey, Biswadip and Chakraborty, Amit},
  journal={Advances in Neural Information Processing Systems},
  volume={34},
  pages={21910--21922},
  year={2021}
}

@InProceedings{neurofluid,
  title = 	 {{N}euro{F}luid: Fluid Dynamics Grounding with Particle-Driven Neural Radiance Fields},
  author =       {Guan, Shanyan and Deng, Huayu and Wang, Yunbo and Yang, Xiaokang},
  booktitle = 	 {Proceedings of the 39th International Conference on Machine Learning},
  pages = 	 {7919--7929},
  year = 	 {2022},
  editor = 	 {Chaudhuri, Kamalika and Jegelka, Stefanie and Song, Le and Szepesvari, Csaba and Niu, Gang and Sabato, Sivan},
  volume = 	 {162},
  series = 	 {Proceedings of Machine Learning Research},
  month = 	 {17--23 Jul},
  publisher =    {PMLR},
  url = 	 {https://proceedings.mlr.press/v162/guan22a.html},
}

@inproceedings{sanchez2020learning,
  title={Learning to simulate complex physics with graph networks},
  author={Sanchez-Gonzalez, Alvaro and Godwin, Jonathan and Pfaff, Tobias and Ying, Rex and Leskovec, Jure and Battaglia, Peter},
  booktitle={International Conference on Machine Learning},
  pages={8459--8468},
  year={2020},
  organization={PMLR}
}

@InProceedings{nclaw,
    title = 	 {Learning Neural Constitutive Laws from Motion Observations for Generalizable {PDE} Dynamics},
    author =       {Ma, Pingchuan and Chen, Peter Yichen and Deng, Bolei and Tenenbaum, Joshua B. and Du, Tao and Gan, Chuang and Matusik, Wojciech},
    booktitle = 	 {Proceedings of the 40th International Conference on Machine Learning},
    pages = 	 {23279--23300},
    year = 	 {2023},
    editor = 	 {Krause, Andreas and Brunskill, Emma and Cho, Kyunghyun and Engelhardt, Barbara and Sabato, Sivan and Scarlett, Jonathan},
    volume = 	 {202},
    series = 	 {Proceedings of Machine Learning Research},
    month = 	 {23--29 Jul},
    publisher =    {PMLR},
    url = 	 {https://proceedings.mlr.press/v202/ma23a.html},
}

@inproceedings{li2023pacnerf,
    title={{PAC}-Ne{RF}: Physics Augmented Continuum Neural Radiance Fields for Geometry-Agnostic System Identification},
    author={Xuan Li and Yi-Ling Qiao and Peter Yichen Chen and Krishna Murthy Jatavallabhula and Ming Lin and Chenfanfu Jiang and Chuang Gan},
    booktitle={The Eleventh International Conference on Learning Representations },
    year={2023},
    url={https://openreview.net/forum?id=tVkrbkz42vc}
}

@inproceedings{murthy2020gradsim,
  title={gradsim: Differentiable simulation for system identification and visuomotor control},
  author={Murthy, J Krishna and Macklin, Miles and Golemo, Florian and Voleti, Vikram and Petrini, Linda and Weiss, Martin and Considine, Breandan and Parent-L{\'e}vesque, J{\'e}r{\^o}me and Xie, Kevin and Erleben, Kenny and others},
  booktitle={International Conference on Learning Representations},
  year={2021}
}

@inproceedings{ma2021risp,
  title={RISP: Rendering-Invariant State Predictor with Differentiable Simulation and Rendering for Cross-Domain Parameter Estimation},
  author={Ma, Pingchuan and Du, Tao and Tenenbaum, Joshua B and Matusik, Wojciech and Gan, Chuang},
  booktitle={International Conference on Learning Representations},
  year={2021}
}

@inproceedings{hudifftaichi,
  title={DiffTaichi: Differentiable Programming for Physical Simulation},
  author={Hu, Yuanming and Anderson, Luke and Li, Tzu-Mao and Sun, Qi and Carr, Nathan and Ragan-Kelley, Jonathan and Durand, Fredo},
  booktitle={International Conference on Learning Representations},
  year={2020}
}

@article{simplicits,
    author = {Modi, Vismay and Sharp, Nicholas and Perel, Or and Sueda, Shinjiro and Levin, David I. W.},
    title = {Simplicits: Mesh-Free, Geometry-Agnostic Elastic Simulation},
    year = {2024},
    issue_date = {July 2024},
    publisher = {Association for Computing Machinery},
    address = {New York, NY, USA},
    volume = {43},
    number = {4},
    issn = {0730-0301},
    url = {https://doi.org/10.1145/3658184},
    doi = {10.1145/3658184},
    journal = {ACM Trans. Graph.},
    month = jul,
    articleno = {117},
    numpages = {11}
}

@inproceedings{vr-gs,
    author = {Jiang, Ying and Yu, Chang and Xie, Tianyi and Li, Xuan and Feng, Yutao and Wang, Huamin and Li, Minchen and Lau, Henry and Gao, Feng and Yang, Yin and Jiang, Chenfanfu},
    title = {VR-GS: A Physical Dynamics-Aware Interactive Gaussian Splatting System in Virtual Reality},
    year = {2024},
    isbn = {9798400705250},
    publisher = {Association for Computing Machinery},
    address = {New York, NY, USA},
    url = {https://doi.org/10.1145/3641519.3657448},
    doi = {10.1145/3641519.3657448},
    booktitle = {ACM SIGGRAPH 2024 Conference Papers},
    articleno = {78},
    numpages = {1},
    location = {Denver, CO, USA},
    series = {SIGGRAPH '24}
}

@article{3dgs,
    author = {Kerbl, Bernhard and Kopanas, Georgios and Leimkuehler, Thomas and Drettakis, George},
    title = {3D Gaussian Splatting for Real-Time Radiance Field Rendering},
    year = {2023},
    issue_date = {August 2023},
    publisher = {Association for Computing Machinery},
    address = {New York, NY, USA},
    volume = {42},
    number = {4},
    issn = {0730-0301},
    url = {https://doi.org/10.1145/3592433},
    doi = {10.1145/3592433},
    journal = {ACM Trans. Graph.},
    month = jul,
    articleno = {139},
    numpages = {14}
}

@article{instantngp,
    author = {M\"{u}ller, Thomas and Evans, Alex and Schied, Christoph and Keller, Alexander},
    title = {Instant neural graphics primitives with a multiresolution hash encoding},
    year = {2022},
    issue_date = {July 2022},
    publisher = {Association for Computing Machinery},
    address = {New York, NY, USA},
    volume = {41},
    number = {4},
    issn = {0730-0301},
    url = {https://doi.org/10.1145/3528223.3530127},
    doi = {10.1145/3528223.3530127},
    journal = {ACM Trans. Graph.},
    month = jul,
    articleno = {102},
    numpages = {15}
}

@article{physics_neural_smoke,
    author = {Chu, Mengyu and Liu, Lingjie and Zheng, Quan and Franz, Aleksandra and Seidel, Hans-Peter and Theobalt, Christian and Zayer, Rhaleb},
    title = {Physics informed neural fields for smoke reconstruction with sparse data},
    year = {2022},
    issue_date = {July 2022},
    publisher = {Association for Computing Machinery},
    address = {New York, NY, USA},
    volume = {41},
    number = {4},
    issn = {0730-0301},
    url = {https://doi.org/10.1145/3528223.3530169},
    doi = {10.1145/3528223.3530169},
    journal = {ACM Trans. Graph.},
    month = jul,
    articleno = {119},
    numpages = {14}
}

@article{geilinger2020add,
  title={Add: Analytically differentiable dynamics for multi-body systems with frictional contact},
  author={Geilinger, Moritz and Hahn, David and Zehnder, Jonas and B{\"a}cher, Moritz and Thomaszewski, Bernhard and Coros, Stelian},
  journal={ACM Transactions on Graphics (TOG)},
  volume={39},
  number={6},
  pages={1--15},
  year={2020},
  publisher={ACM New York, NY, USA}
}

@article{jiang2016material,
  title={The material point method for simulating continuum materials},
  author={Jiang, Chenfanfu and Schroeder, Craig and Teran, Joseph and Stomakhin, Alexey and Selle, Andrew},
  journal = {ACM SIGGRAPH 2016 Courses},
  pages={1--52},
  year={2016}
}

@article{huang2024dreamphysics,
  title={DreamPhysics: Learning Physical Properties of Dynamic 3D Gaussians with Video Diffusion Priors},
  author={Huang, Tianyu and Zeng, Yihan and Li, Hui and Zuo, Wangmeng and Lau, Rynson WH},
  journal={arXiv preprint arXiv:2406.01476},
  year={2024}
}

@inproceedings{
    abou-chakra2024physically,
    title={Physically Embodied Gaussian Splatting: A Realtime Correctable World Model for Robotics},
    author={Jad Abou-Chakra and Krishan Rana and Feras Dayoub and Niko Suenderhauf},
    booktitle={8th Annual Conference on Robot Learning},
    year={2024},
    url={https://openreview.net/forum?id=AEq0onGrN2}
}

@article{
    jiang2025phystwin,
    title={PhysTwin: Physics-Informed Reconstruction and Simulation of Deformable Objects from Videos},
    author={Jiang, Hanxiao and Hsu, Hao-Yu and Zhang, Kaifeng and Yu, Hsin-Ni and Wang, Shenlong and Li, Yunzhu},
    journal={arXiv preprint arXiv:2503.17973},
    year={2025}
}

@misc{li2024robogsimreal2sim2realroboticgaussian,
        title={RoboGSim: A Real2Sim2Real Robotic Gaussian Splatting Simulator}, 
        author={Xinhai Li and Jialin Li and Ziheng Zhang and Rui Zhang and Fan Jia and Tiancai Wang and Haoqiang Fan and Kuo-Kun Tseng and Ruiping Wang},
        year={2024},
        eprint={2411.11839},
        archivePrefix={arXiv},
        primaryClass={cs.RO},
        url={https://arxiv.org/abs/2411.11839}, 
  }

@article{xu2019densephysnet,
  title={Densephysnet: Learning dense physical object representations via multi-step dynamic interactions},
  author={Xu, Zhenjia and Wu, Jiajun and Zeng, Andy and Tenenbaum, Joshua B and Song, Shuran},
  journal={arXiv preprint arXiv:1906.03853},
  year={2019}
}
\bibliographystyle{tmlr}

\appendix

\section{Optimization Trajectory Analysis}

\begin{figure}[ht]
    \centering
    \includegraphics[width=0.75\linewidth]{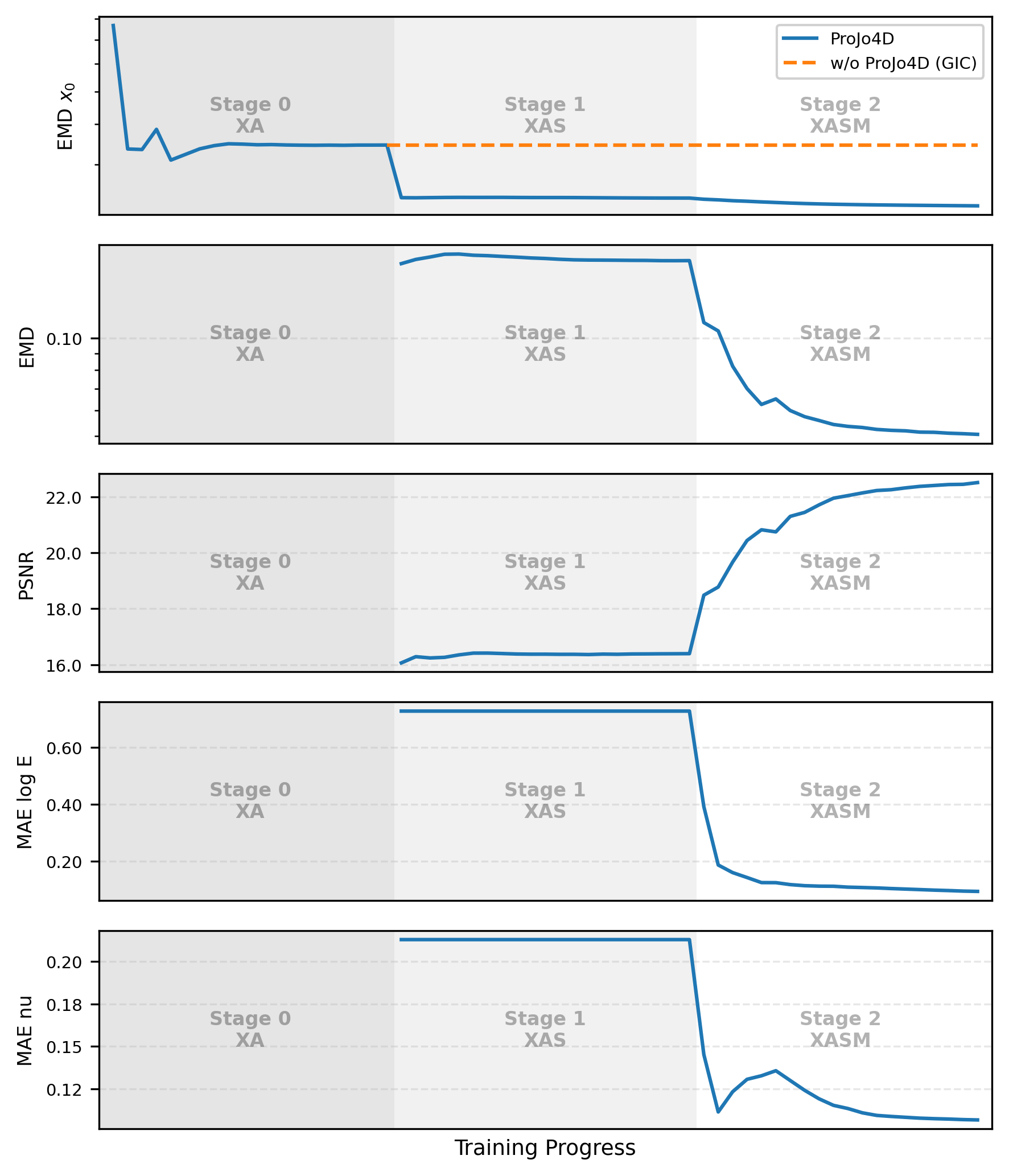}
    \caption{
        \textbf{Average optimization trajectories for Spring-Gaus scenes.}
        Background shading denotes Stage 0 (dark gray), Stage 1 (light gray), and Stage 2 (white).
        EMD and PSNR evaluate future-state prediction; MAE measures material parameter accuracy.
        EMD$_{x_0}$ indicates error in canonical space. Note: Stage 0 is rescaled for visualization due to its higher iteration count.
        By introducing physics-informed gradients during joint optimization, ProJo4D refines geometry more effectively than existing methods.
    }
    \label{fig:trajectories}
\end{figure}

Fig.~\ref{fig:trajectories} provides an in-depth analysis of the optimization dynamics across different stages of ProJo4D. This figure visualizes how various metrics, including Earth Mover's Distance (EMD), Peak Signal-to-Noise Ratio (PSNR), and Mean Absolute Error (MAE) for material parameters, evolve through Stage 0 (initial 4D representation learning), Stage 1 (joint optimization of positions, appearance, and velocity), and Stage 2 (full joint optimization including material parameters).

The trajectory analysis demonstrates that by introducing physics-informed gradients during joint optimization, ProJo4D refines geometry more effectively than existing sequential methods. The joint optimization of positions alongside physical parameters in Stages 1 and 2 improves the quality of the initial point clouds, which in turn improves physical parameter estimation.

\begin{table*}[ht!]
    \caption{\textbf{Ablation on Stage 1 design choices across different material types.} We evaluate 3D future state prediction accuracy (CD) and initial velocity estimation (MAE $v_0$) on the PAC-NeRF dataset. Each row represents a different Stage 1 configuration, where X, A, S, and M denote positions, appearances, velocity, and material parameters, respectively. Bold values indicate best performance.}
    \label{tab:ablation}
    \centering
    \resizebox{\linewidth}{!}{
        \begin{tabular}{rlrrrrr}
            \toprule
             & Stage 1 & \textbf{Elastic} & \textbf{Newtonian} & \textbf{Non-Newtonian} & \textbf{Plasticine} & \textbf{Sand} \\
            \midrule
            % ------------------------- CD -------------------------
            \multirow{6}{*}{CD \(\downarrow\)}
            & S & 0.953 $\pm$ 0.295 & 6.319 $\pm$ 7.854 & 9.668 $\pm$ 4.543 & 21.891 $\pm$ 17.848 & 2.727 $\pm$ 0.531 \\
            & M & 1.020 $\pm$ 0.314 & 4.830 $\pm$ 6.432 & 9.205 $\pm$ 3.559 & 20.470 $\pm$ 17.660 & 4.067 $\pm$ 2.469 \\
            & SM & 1.057 $\pm$ 0.349 & 4.896 $\pm$ 4.254 & 9.682 $\pm$ 4.126 & 20.434 $\pm$ 17.976 & 2.743 $\pm$ 0.543 \\
            & \textbf{XAS (ProJo4D)} & \textbf{0.913 $\pm$ 0.301} & \textbf{0.339 $\pm$ 0.108} & \textbf{0.473 $\pm$ 0.248} & \textbf{1.103 $\pm$ 0.948} & \textbf{0.264 $\pm$ 0.017} \\
            & XAM & 1.053 $\pm$ 0.339 & 3.226 $\pm$ 1.607 & 9.164 $\pm$ 4.587 & 20.368 $\pm$ 16.643 & 2.457 $\pm$ 0.359 \\
            & XASM (Full Joint) & 1.318 $\pm$ 1.117 & 0.346 $\pm$ 0.095 & 8.104 $\pm$ 13.563 & 17.678 $\pm$ 18.170 & 53.564 $\pm$ 19.404 \\
            \midrule
            % ------------------------- MAE v0 -------------------------
            \multirow{6}{*}{MAE $v_0$ \(\downarrow\)}
            & S & \textbf{0.007 $\pm$ 0.003} & 0.074 $\pm$ 0.045 & 0.132 $\pm$ 0.031 & 0.102 $\pm$ 0.091 & 0.085 $\pm$ 0.039 \\
            & M & 0.035 $\pm$ 0.025 & 0.089 $\pm$ 0.045 & 0.154 $\pm$ 0.052 & 0.131 $\pm$ 0.107 & 0.150 $\pm$ 0.128 \\
            & SM & 0.013 $\pm$ 0.009 & 0.073 $\pm$ 0.044 & 0.149 $\pm$ 0.086 & 0.128 $\pm$ 0.070 & 0.095 $\pm$ 0.034 \\
            & \textbf{XAS (ProJo4D)} & \textbf{0.007 $\pm$ 0.003} & \textbf{0.008 $\pm$ 0.002} & \textbf{0.005 $\pm$ 0.003} & \textbf{0.024 $\pm$ 0.056} & \textbf{0.005 $\pm$ 0.003} \\
            & XAM & 0.099 $\pm$ 0.036 & 0.125 $\pm$ 0.045 & 0.162 $\pm$ 0.031 & 0.173 $\pm$ 0.091 & 0.252 $\pm$ 0.056 \\
            & XASM (Full Joint) & 0.020 $\pm$ 0.033 & 0.008 $\pm$ 0.004 & 0.080 $\pm$ 0.099 & 0.092 $\pm$ 0.102 & 0.046 $\pm$ 0.032 \\
            \bottomrule
        \end{tabular}
    }
\end{table*}

\section{Stage 1 Design Choices}
\label{sec:abl_stage1_design}
In the main paper, we compare ProJo4D against alternative optimization strategies (sequential, cyclic).
Here, we provide additional analysis on the design choices within Stage 1 of our progressive optimization pipeline.
Specifically, we investigate which parameters to optimize in Stage 1 before transitioning to full joint optimization in Stage 2.

We conduct experiments on the PAC-NeRF dataset, which contains multiple material models.
Because different material models involve distinct parameterizations, we report only metrics common across all models: Chamfer distance (CD) for future dynamics estimation and mean absolute error (MAE) for initial velocity.
Tab.~\ref{tab:ablation} compares different Stage 1 configurations: optimizing only velocity (S), only material parameters (M), both sequentially (SM), velocity jointly with positions and appearance (XAS, our default), material jointly with positions and appearance (XAM), and all parameters jointly (XASM, i.e., full joint optimization immediately after Stage 0).

Our progressive method (XAS) demonstrates stability across different material models.
Full joint optimization (XASM) achieves comparable performance to our method for relatively simple material models (Elastic and Newtonian), but shows significant degradation for more complex materials (Non-Newtonian, Plasticine, and Sand).
This confirms that while full joint optimization can work for simpler models, it becomes unreliable as material complexity increases.

The results also show that optimizing velocity first (XAS) consistently outperforms material-first (XAM) across all materials.
Initial velocity directly affects early-frame dynamics, making it easier to constrain from the first few frames, whereas material parameters govern longer-term deformation behavior that requires more frames to disambiguate.

\section{Hyperparameters}
\label{ssec:hyperparameters}

\textbf{Stage 0: 3D/4D Representation Learning.}
To ensure a fair comparison, ProJo4D and the GIC baseline share identical 4D representations. Both are obtained after 40K iterations: a 3K-iteration warmup for static initialization followed by 37K iterations using deformation networks.

\textbf{Stages 1 \& 2: Joint Optimization.}
We adopt the same hyperparameters as GIC to isolate the performance gains from our progressive joint strategy.

\textbf{Spring-Gaus Dataset.}
ProJo4D uses 100 iterations each for Stage 1 and Stage 2. While GIC requires a 30K-iteration ``Stage 3'' for appearance refinement, ProJo4D skips this because appearance is already optimized jointly during Stages 1 and 2.

\textbf{PAC-NeRF Dataset.}
We maintain GIC's iteration schedule per material:
\begin{itemize}
    \item \textbf{Elastic/Sand}: 100 for Stage 1, 150 for Stage 2
    \item \textbf{Newtonian}: 100 for Stage 1, 250 for Stage 2
    \item \textbf{Non-Newtonian}: 100 for Stage 1, 350 for Stage 2
    \item \textbf{Plasticine}: 100 for Stage 1, 300 for Stage 2
\end{itemize}

\begin{table}
    \centering
    \caption{
        \textbf{Training configurations for Tab.~\ref{tab:abl_alts}.} 
        Batch sizes are in parentheses. For multi-view/multi-frame stages, we use: (number of cameras $\times$ number of frames).
    }
    \label{tab:abl_alts_details}
    \begin{tabular}{rlll}
        \toprule
         & ProJo4D & Sequential (GIC) & Sequential+ \\
        \midrule
        Stage 0 & 40K (1) & 40K (1) & 47K (1) \\
        Stage 1 & 100 (3$\times$3) & 100 (3$\times$3) & 200 (3$\times$3) \\
        Stage 2 & 100 (3$\times$20) & 100 (3$\times$20) & 100 (3$\times$20) \\
        Stage 3 & 0 & 40K (1) & 40K (1) \\
        \bottomrule
    \end{tabular}
\end{table}

\section{Experimental Details for the GSO Dataset}

While Vid2Sim, PAC-NeRF, and GIC all utilize the Neo-Hookean model for elastic objects, their stress formulations vary.

\textbf{Stress Formulation Variants.}
PAC-NeRF and GIC define the Kirchhoff stress tensor $\tau$ as:
\begin{equation}
    \tau_{PAC-NeRF} = \mu FF^T + (\lambda J - \mu) I,
\end{equation}
where $F$ is the deformation gradient, $J = \det(F)$, and $\mu, \lambda$ are Lamé parameters.
Conversely, Vid2Sim (following Simplicits) uses:
\begin{equation}
    \tau_{Simplicits} = \mu FF^T + (\lambda (J-1) - \mu) J I.
    \label{eq:simplicits}
\end{equation}

\textbf{Fair Comparison.}
For a controlled comparison on the GSO dataset, we reran both GIC and ProJo4D using the Simplicits formulation (Eq.~\ref{eq:simplicits}).

\end{document}